\documentclass[10pt,twocolumn,letterpaper]{article}
\usepackage[T1]{fontenc}
\usepackage{preprint}

\usepackage{microtype}

\let\originaleqref\eqref

\usepackage{amsmath,amsfonts,bm}

\def\eqref#1{equation~\ref{#1}}

\def\1{\bm{1}}

\DeclareMathAlphabet{\mathsfit}{\encodingdefault}{\sfdefault}{m}{sl}
\SetMathAlphabet{\mathsfit}{bold}{\encodingdefault}{\sfdefault}{bx}{n}

\let\eqref\originaleqref
\usepackage{amsthm}
\usepackage{float}
\usepackage{array}

\newtheorem{theorem}{Theorem}

\newcommand{\experimentplot}[4]{%
  \begin{minipage}[c]{#2}\centering
    \IfFileExists{#1}{\includegraphics[width=\linewidth]{#1}}{%
      \fbox{\parbox[c][#3][c]{\dimexpr\linewidth-2\fboxsep-2\fboxrule\relax}{%
        \centering\footnotesize\textbf{Figure pending}\par\smallskip #4}}%
    }%
  \end{minipage}%
}

\newcommand{\vla}{f_{\omega}}
\newcommand{\Achunk}{\mathbf A}                      %
\newcommand{\Ares}{\widetilde{\mathbf A}}            %
\newcommand{\Ahat}{\widehat{\mathbf A}}              %
\newcommand{\Vprof}{\mathbf V}                       %
\newcommand{\Dchunk}{\mathbf D}                      %
\newcommand{\Dmean}{\bar{\mathbf D}}                 %
\newcommand{\Dcen}{\boldsymbol\Delta}                %
\newcommand{\Dnorm}{\boldsymbol\Delta_{\mathcal N}}
\newcommand{\obs}{\mathbf o}                         %
\newcommand{\noise}{\boldsymbol\epsilon}             %
\newcommand{\sat}{\operatorname{sat}}

\usepackage{tikz}
\usetikzlibrary{arrows.meta,calc,positioning}
\newif\ifshownotes
\shownotestrue
\ifdefined\hidenotes\shownotesfalse\fi
\definecolor{ambigcolor}{rgb}{0.72,0.08,0.52}
\definecolor{missingcolor}{rgb}{0.80,0.10,0.08}
\newcounter{ambigctr}
\newcounter{missingctr}

\definecolor{polishcolor}{rgb}{0.04,0.38,0.56}
\newcounter{polishctr}

\newif\ifpendingstudies
\pendingstudiestrue
\ifdefined\measuredonly\pendingstudiesfalse\fi

\usepackage{refcount}
\usepackage{pdftexcmds}
\makeatletter
\newcommand{\xr@list@main}{}
\newcommand{\xr@list@supp}{}
\newcommand{\xrdeclare}[3]{%
  \expandafter\gdef\csname xr@#1@#2\endcsname{#3}%
  \expandafter\g@addto@macro\csname xr@list@#1\endcsname{\xr@do{#2}}}
\newcommand{\xr@get}[2]{%
  \@ifundefined{xr@#1@#2}{%
    \PackageError{xref}{Cross-document label '#2' is not declared in
      xref_table.tex}{Add an \string\xrdeclare\space line for it.}%
    \textbf{??}}{\csname xr@#1@#2\endcsname}}
\newcommand{\suppref}[1]{\xr@get{supp}{#1}}
\newcommand{\mainref}[1]{\xr@get{main}{#1}}
\newcommand{\xrverify}[1]{%
  \begingroup
  \def\xr@do##1{%
    \@ifundefined{r@##1}{%
      \PackageWarningNoLine{xref}{Label '##1' is declared in xref_table.tex
        but is not defined in this document (yet); rerun LaTeX}}{%
      \edef\xr@actual{\getrefnumber{##1}}%
      \edef\xr@stored{\csname xr@#1@##1\endcsname}%
      \ifnum\pdf@strcmp{\xr@actual}{\xr@stored}=0 \else
        \PackageError{xref}{Stale cross-document number: label '##1' is now
          '\xr@actual' but xref_table.tex says '\xr@stored'}{Update the
          \string\xrdeclare\space line for '##1' in xref_table.tex, then
          recompile main.tex and supplementary.tex.}%
      \fi}}%
  \csname xr@list@#1\endcsname
  \endgroup}
\makeatother
\xrdeclare{supp}{supp:alg}{A}
\xrdeclare{supp}{supp:alg_vi}{A.3}
\xrdeclare{supp}{supp:details}{B}
\xrdeclare{supp}{supp:conventions}{B.1}
\xrdeclare{supp}{supp:forecast}{B.2}
\xrdeclare{supp}{supp:centering}{B.3}
\xrdeclare{supp}{supp:conditioning}{B.4}
\xrdeclare{supp}{supp:analysis}{C}
\xrdeclare{supp}{supp:ablation}{D}
\xrdeclare{supp}{supp:components}{D.1}
\xrdeclare{supp}{supp:noise}{D.2}
\xrdeclare{supp}{supp:tracking}{E}
\xrdeclare{supp}{supp:runtime}{G}
\xrdeclare{supp}{supp:limitations}{H}
\xrdeclare{main}{sec:method}{3}
\xrdeclare{main}{subsec:kinematic_projection}{3.1}
\xrdeclare{main}{subsec:residual_target}{3.2}
\xrdeclare{main}{subsec:inference}{3.3}
\xrdeclare{main}{subsec:residual_distribution}{3.4}
\xrdeclare{main}{subsec:velocity_innovation}{3.5}
\xrdeclare{main}{sec:exp_setup}{4.1}
\xrdeclare{main}{sec:exp_tracking}{4.2}
\xrdeclare{main}{sec:exp_dom}{4.3}
\xrdeclare{main}{sec:exp_real}{4.4}
\xrdeclare{main}{sec:exp_ablations}{4.5}
\xrdeclare{main}{tab:dom_main}{2}
\xrdeclare{main}{tab:tracking}{1}
\xrdeclare{main}{tab:real_robot_results}{3}
\xrdeclare{main}{tab:innovation_ablation}{4}
\xrdeclare{main}{tab:velocity_uncertainty}{5}
\xrdeclare{main}{fig:framework_overview}{1}
\xrdeclare{main}{fig:kra_pipeline}{2}
\xrdeclare{main}{fig:velocity_innovation}{3}
\xrdeclare{main}{fig:demonstration_sequence}{4}
\xrdeclare{main}{fig:real_policy_sequence}{5}
\xrdeclare{main}{eq:continuous_velocity_forecast}{1}
\xrdeclare{main}{eq:velocity_forecast}{2}
\xrdeclare{main}{eq:forecast_displacement}{3}
\xrdeclare{main}{eq:kinematic_projection}{4}
\xrdeclare{main}{eq:displacement_chunk}{5}
\xrdeclare{main}{eq:centered_displacement}{6}
\xrdeclare{main}{eq:residual_target}{7}
\xrdeclare{main}{eq:residual_path}{8}
\xrdeclare{main}{eq:krfm_loss}{9}
\xrdeclare{main}{eq:residual_euler}{10}
\xrdeclare{main}{eq:residual_to_raw_inference}{11}
\xrdeclare{main}{eq:flow_field_correspondence}{12}
\xrdeclare{main}{eq:covariance_difference}{13}
\xrdeclare{main}{eq:velocity_innovation}{14}
\xrdeclare{main}{eq:innovation_correction}{15}
\xrdeclare{main}{eq:innovation_action}{16}

\usepackage{algorithm}
\usepackage{algpseudocode}

\definecolor{linkblue}{rgb}{0.21,0.49,0.74}
\usepackage[pagebackref,breaklinks,colorlinks,allcolors=linkblue]{hyperref}
\hypersetup{pdfauthor={Aditya Parameshwaran, Nianyi Li, Yue Wang},pdftitle={Adapting Vision-Language-Action Policies to Moving Objects with Kinematic Residuals}}

\renewcommand{\suppref}[1]{\ref{#1}}
\renewcommand{\mainref}[1]{\ref{#1}}
\newcommand{\meq}[1]{Eq.~(\ref{#1})}

\title{Adapting Vision-Language-Action Policies to Moving Objects with \\
 Kinematic Residuals}

\author{Aditya Parameshwaran \quad Nianyi Li \quad Yue Wang\\
Clemson University\\
{\tt\small aparame@clemson.edu, nianyil@clemson.edu, yue6@clemson.edu,}}

\newsavebox{\paperfigurebox}
\newcommand{\checkpaperfigure}[1]{\usebox{\paperfigurebox}}
\makeatletter
\apptocmd{\@maketitle}{\par\noindent
\begin{lrbox}{\paperfigurebox}\begin{minipage}{\textwidth}
  \centering
  \captionsetup{skip=3pt,hypcap=false}
  \includegraphics[width=0.8\textwidth,height=.23\textheight,keepaspectratio]{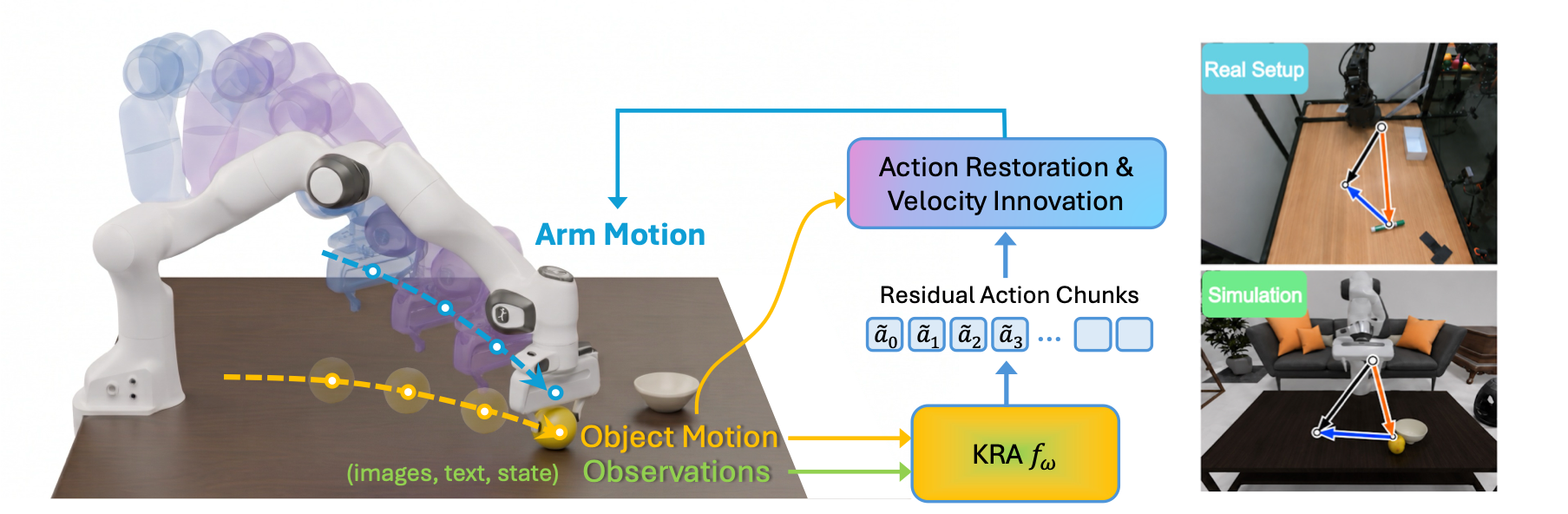}
  \captionof{figure}{\textbf{Kinematic Residual Adaptation for moving-object manipulation.}
  KRA subtracts planning-time velocity-profile displacement (VPD) to learn residual actions and restores it after sampling. During execution, velocity innovation (VI) adds a bounded correction for observed departures from the forecast.}
  \label{fig:framework_overview}
\end{minipage}\end{lrbox}%
\checkpaperfigure{overview}\par\medskip
}{}{
  \PackageError{paper}{Could not attach overview to title block}{}
}
\makeatother
\begin{document}
\maketitle

\begin{abstract}

Vision-language-action (VLA) policies are developed mostly in quasi-static scenes and degrade sharply once the target object moves. The action chunks that realize the same grasp differ for every object velocity, and standard fine-tuning leaves the policy to learn this dependence from limited motion diversity.
We propose Kinematic Residual Adaptation (KRA), which removes the predictable part of this dependence from the learning problem.
KRA integrates a planning-time constant-acceleration forecast into object displacement over the action horizon and subtracts it from the translational targets of each demonstrated chunk. It then fine-tunes a pretrained flow-based VLA on the residual chunks and restores the same displacement after sampling.
Training and generation are linked by an exact, invertible change of action coordinates that keeps the backbone's flow-matching objective and sampler. During execution, bounded velocity innovation adds a displacement-error correction without re-querying the policy.
With simulator-state velocity, KRA raises the average success of a velocity-conditioned DynamicVLA on a low-clutter variant of the Dynamic Object Manipulation benchmark from 25.9\% to 30.8\%, improves three other VLA backbones, and increases moving-target tracking coverage at training and held-out speeds.

\end{abstract}

\section{Introduction}
\label{sec:intro}

Vision-language-action (VLA) policies couple pretrained vision-language
representations with action generation and perform a broad range of
language-conditioned manipulation
tasks~\citep{brohan2023rt1,zitkovich2023rt2,kim2024openvla,black2025pi0,physicalintelligence2025pi05}.
Most of this progress concerns quasi-static scenes. When the target moves,
as on a conveyor or during a handover, a correct grasp prediction can still
fail because the object is no longer where the action chunk assumed it would
be: on the Dynamic Object Manipulation (DOM) benchmark, general-purpose VLAs
reach average success rates below 14\%~\citep{xie2026dynamicvla}.

Two problems are entangled in this failure. The first is temporal:
observations become stale while a chunk is inferred and executed, which
recent systems mitigate with compact backbones, continuous inference, action
streaming, and future-state
conditioning~\citep{xie2026dynamicvla,tang2025vlash,black2026real,shukor2025smolvla}.
The second persists even with zero latency and concerns \emph{what the
policy is asked to predict}. For the same instruction, object, and intended
grasp, the end-effector positions in a demonstrated chunk shift with the
displacement that the object accumulates over the action horizon. A policy
fine-tuned on raw demonstrations must learn this velocity-dependent offset
jointly with the manipulation behavior. Feeding velocity to the network
makes the signal available, but the mapping from velocity to offset is still
learned from demonstrations that cover a limited range of speeds and
directions.

Our starting point is that \textit{the dominant part of this dependence does
not need to be learned: to first order, it is kinematics.} Over a short
horizon, the displacement of the target is the time integral of its
velocity, and the translational part of the demonstrated chunk moves with
it. Once this displacement is removed, the remaining action describes the
manipulation in a frame that travels with the object.

We introduce \textbf{Kinematic Residual Adaptation (KRA)}, a fine-tuning
framework built on this observation (Figures~\ref{fig:framework_overview}
and~\ref{fig:kra_pipeline}). At planning time, KRA forecasts the object
velocity over the action horizon with a constant-acceleration model and
integrates it into a velocity-profile displacement (VPD). This displacement
is subtracted from the translational targets of each demonstrated chunk, the
flow-based VLA is fine-tuned on the resulting \emph{residual chunks} with its
original flow-matching objective, and the same displacement is added back
after sampling. Unlike compensation applied only at inference, the policy is
trained in the coordinates in which it samples; unlike velocity conditioning
alone, the targets themselves are transformed. Velocity remains a network
input, so approach and contact strategies can still depend on motion.
Because a chunk is executed open loop between replanning steps, a bounded,
rate-limited \emph{velocity innovation} (VI) additionally corrects the
commanded translation with the accumulated difference between observed and
forecast object motion.

Our contributions are:
\begin{itemize}
  \item \textbf{A kinematic residual action parameterization} that factors
  the forecast object displacement out of demonstrated chunks and restores
  it after sampling. The transformation is exact and invertible and leaves
  the objective, sampler, and number of network evaluations of the backbone
  unchanged.
  \item \textbf{Execution-time velocity innovation} that compensates
  forecast errors within a chunk without querying the policy again.
  \item \textbf{Controlled evaluation.} In hover tracking, a $2{\times}2$
  ablation separates residualization from neural velocity input and tests
  held-out speeds. On nine low-clutter DOM conditions, KRA raises the
  success of a velocity-conditioned DynamicVLA~\citep{xie2026dynamicvla}
  from 25.89\% to 30.78\% and improves
  SmolVLA~\citep{shukor2025smolvla}, VLASH~\citep{tang2025vlash}, and
  $\pi_{0.5}$~\citep{physicalintelligence2025pi05}; a physical
  pick-and-place study uses visually estimated velocity.
\end{itemize}

\section{Related Work}
\label{sec:related}

\noindent\textbf{Vision-language-action policies.}
RT-1, RT-2, Octo, and OpenVLA map images and instructions to robot actions
with transformers trained on large robot
datasets~\citep{brohan2023rt1,zitkovich2023rt2,octo2024,kim2024openvla,openx2024}.
$\pi_0$, $\pi_{0.5}$, SmolVLA, and GR00T-N1 add diffusion or flow-matching
action heads, compact backbones, and heterogeneous
data~\citep{black2025pi0,physicalintelligence2025pi05,shukor2025smolvla,bjorck2025gr00t},
and can be fine-tuned efficiently for new tasks~\citep{kim2025openvlaoft}.
These models are trained and evaluated mostly on quasi-static tasks.

\vspace{3pt}
\noindent\textbf{Dynamic manipulation and real-time execution.}
Classical approaches to moving targets combine explicit motion prediction
with planning or servoing, as in visual
servoing~\citep{chaumette2006visual}, catching~\citep{kim2014catching}, and
dynamic grasping and
handover~\citep{akinola2021dynamic,liu2023targetreferenced,fang2023anygrasp,wang2024genh2r}.
For learned policies, recent work concentrates on the delay between
observation and action: real-time chunking generates the next chunk
asynchronously~\citep{black2026real,black2025trainingrtc}, and related
schemes resample, stream, or correct chunks with the latest
observation~\citep{liu2025bid,sendai2025a2c2,lu2026faster,park2026pir2,chen2026ravel,dong2026realtimerl}.
VLASH conditions inference on the future robot
state~\citep{tang2025vlash}, SmolVLA decouples inference from
execution~\citep{shukor2025smolvla}, and DynamicVLA combines a compact model
with continuous inference and latency-aware action streaming and introduces
the DOM benchmark~\citep{xie2026dynamicvla}. Concurrent dynamic-manipulation
VLAs add motion cues, memory, or learned future
prediction~\citep{syed2026ahead,ma2026foretime,ye2026d2vla,li2026dsdynvla,li2026reactive}.
These methods determine \emph{when} actions are computed and executed, or
learn to anticipate motion inside the network. KRA addresses \emph{what} is
learned: it brings the kinematic prediction of classical pipelines into the
supervision of a VLA and is compatible with these execution schedules. Its
velocity input can be estimated causally by segmentation, point, or pose
tracking~\citep{ravi2024sam2,xiong2025efficient,karaev2024cotracker,wen2024foundationpose}.

\vspace{3pt}
\noindent\textbf{Action generation and representations.}
Diffusion Policy~\citep{chi2023diffusionpolicy} and flow
matching~\citep{lipman2023flow,tong2024improving,liu2023rectified} generate
chunks of future actions~\citep{zhao2023act} in recent
VLAs~\citep{black2025pi0}. The action representation itself affects
learning: frequency-space tokenization~\citep{pertsch2025fast},
relative-trajectory actions~\citep{chi2024umi}, and object-centric pose or
point-track targets~\citep{hsu2025spot,wen2024atm}. Residual policy learning
adds a learned correction to a controller or a frozen imitation or VLA
policy~\citep{johannink2019residual,silver2018residual,ankile2024resip,yuan2025policydecorator,ankile2025resfit,xiao2026pld,kim2026objectcentricresidual}.
KRA instead fine-tunes the VLA itself on demonstration actions from which
the kinematic displacement has been removed, and restores that displacement
after sampling.

\section{Kinematic Residual Adaptation}
\label{sec:method}

KRA adapts a pretrained flow-based VLA to moving objects by changing the coordinates in which its action chunks are supervised and generated (Figure~\ref{fig:kra_pipeline}). A \emph{motion projection} converts the observed object velocity into a displacement of the action chunk (Sec.~\ref{subsec:kinematic_projection}). The VLA is fine-tuned
on chunks from which this displacement has been removed
(Sec.~\ref{subsec:residual_target}), and the displacement is added back after sampling (Sec.~\ref{subsec:inference}). A \emph{velocity innovation} corrects forecast errors during execution
(Sec.~\ref{subsec:velocity_innovation}).

\providecommand{\KRAeq}[1]{\ref{#1}}
\begin{figure*}[t]
\centering
\resizebox{0.9\textwidth}{!}{%
\begin{tikzpicture}[
    x=1cm, y=1cm, font=\small,
    >={Stealth[length=1.9mm,width=1.5mm]},
    flow/.style={->, line width=0.6pt, draw=black, rounded corners=0pt},
    grad/.style={->, line width=0.6pt, draw=kraBlue, dashed},
    panel/.style={draw=black!28, line width=0.7pt, rounded corners=3.5pt},
    blk/.style={draw=black!50, fill=black!6, line width=0.6pt, rounded corners=2pt,
                minimum height=0.68cm, inner xsep=5pt, inner ysep=2pt, align=center},
    trainblk/.style={blk, draw=kraBlue, fill=kraBlue!14},
    op/.style={circle, draw=black, line width=0.6pt, inner sep=0pt, minimum size=3.5mm,
               fill=white},
    lbl/.style={inner sep=1.5pt},
    ttl/.style={anchor=west, inner sep=0pt},
    fire/.pic={
      \begin{scope}[scale=0.135]
        \fill[kraFlame]
          (0,-1) .. controls (-0.95,-1) and (-1.05,-0.05) .. (-0.5,0.4)
                 .. controls (-0.5,0.05) and (-0.25,0.0) .. (-0.2,0.25)
                 .. controls (-0.3,0.7) and (-0.05,0.95) .. (0.2,1.15)
                 .. controls (0.15,0.65) and (0.95,0.35) .. (0.85,-0.3)
                 .. controls (0.8,-0.8) and (0.45,-1) .. (0,-1) -- cycle;
        \fill[kraFlameIn]
          (0.02,-0.95) .. controls (-0.5,-0.95) and (-0.5,-0.35) .. (0.0,0.15)
                 .. controls (0.5,-0.35) and (0.52,-0.95) .. (0.02,-0.95) -- cycle;
      \end{scope}},
    snow/.pic={
      \begin{scope}[scale=0.135, line width=0.75pt, draw=kraSnow, line cap=round]
        \foreach \a in {90,150,...,390}{
          \draw (0,0) -- (\a:1.05);
          \draw (\a:0.62) -- ++(\a+50:0.36);
          \draw (\a:0.62) -- ++(\a-50:0.36);
        }
      \end{scope}},
  ]
  \definecolor{kraBlue}{RGB}{45,85,165}
  \definecolor{kraFlame}{RGB}{238,90,36}
  \definecolor{kraFlameIn}{RGB}{255,196,48}
  \definecolor{kraSnow}{RGB}{86,170,232}

  \node[draw=black!50, fill=black!6, line width=0.6pt, minimum width=17.4cm,
        minimum height=0.6cm, inner sep=0pt, anchor=south west] (bar) at (0,0)
        {Constant-acceleration projection:\quad
         $(\mathbf v(t_0),\hat{\mathbf a}_{\chi})\;\mapsto\;\Vprof
           \;\mapsto\;\Dcen=\Dchunk(\Vprof)-\Dmean$
         \quad(Eqs.~\KRAeq{eq:velocity_forecast}--\KRAeq{eq:centered_displacement})};

  \draw[panel] (0,1.3)    rectangle (5.8,5.2);
  \draw[panel] (6.1,1.3)  rectangle (11.6,5.2);
  \draw[panel] (11.9,1.3) rectangle (17.4,5.2);
  \node[ttl] at (0.18,4.9)  {\textbf{(1)} Residual fine-tuning};
  \node[ttl] at (6.28,4.9)  {\textbf{(2)} Generation and restoration};
  \node[ttl] at (12.08,4.9) {\textbf{(3)} Execution with innovation};

  \node[op] (sub) at (1.1,2.7) {};
  \draw[line width=0.6pt] ($(sub)+(-0.1,0)$) -- ($(sub)+(0.1,0)$);
  \node[lbl] (x1) at (0.38,2.7) {$\Achunk$};
  \draw[flow] (x1) -- (sub);
  \draw[flow] (1.1,0.6) -- (sub.south);
  \node[lbl, anchor=west] at (1.15,0.95) {$\Dcen$};

  \node[blk] (path) at (2.3,2.7) {$\mathcal N$\\FM path};
  \draw[flow] (sub) -- (path);
  \node[lbl] (noise) at (2.3,1.68) {$\noise,\,s$};
  \draw[flow] (noise) -- (path);

  \node[trainblk, minimum width=1.3cm] (f1) at (4.5,2.7) {$\vla$\hspace{0.42cm}};
  \pic at ($(f1.center)+(0.3,0.0)$) {fire};
  \draw[flow] (path) -- node[lbl, above] {$\widetilde{\mathbf X}^{s}$} (f1);
  \node[lbl] (z1) at (4.5,1.68) {$\mathbf z$};
  \draw[flow] (z1) -- (f1);

  \node[lbl, text=kraBlue] (L) at (4.1,4.25)
     {$\mathcal L_{\mathrm{KRA}}$ {\footnotesize(Eq.~\KRAeq{eq:krfm_loss})}};
  \draw[flow] (sub.north) -- (1.1,4.25) -- (L.west);
  \node[lbl, anchor=west] at (1.15,3.5) {$\mathcal N(\Ares)$ {\footnotesize(Eq.~\KRAeq{eq:residual_target})}};
  \draw[flow] (f1.east) -- (5.55,2.7) -- (5.55,4.25) -- (L.east);
  \draw[grad] (L.south -| f1.north) -- (f1.north)
     node[lbl, midway, left, text=kraBlue] {$\nabla_{\!\omega}$};

  \node[blk, minimum width=1.3cm] (f2) at (8.2,3.7) {$\vla$\hspace{0.42cm}};
  \pic at ($(f2.center)+(0.3,0.0)$) {snow};
  \node[lbl] (n2) at (6.72,3.7) {$\widetilde{\mathbf X}^{0}$};
  \node[lbl] at (6.86,3.28) {\footnotesize noise};
  \draw[flow] (n2) -- (f2);
  \node[lbl] (z2) at (8.2,2.55) {$\mathbf z$};
  \draw[flow] (z2) -- (f2);

  \node[lbl] (st) at (10.15,3.7) {$\widetilde{\mathbf X}^{s_k}$};
  \draw[flow] ($(f2.east)+(0,0.1)$) -- ($(st.west)+(0,0.1)$);
  \draw[flow] ($(st.west)+(0,-0.1)$) -- ($(f2.east)+(0,-0.1)$);
  \node[lbl] at (9.4,4.2) {\footnotesize $\times K$ (Eq.~\KRAeq{eq:residual_euler})};

  \node[op] (add2) at (10.15,2.1) {};
  \draw[line width=0.6pt] ($(add2)+(-0.1,0)$) -- ($(add2)+(0.1,0)$)
                          ($(add2)+(0,-0.1)$) -- ($(add2)+(0,0.1)$);
  \draw[flow] (st) -- node[lbl, right] {$\mathcal N^{-1}(\widetilde{\mathbf X}^{1})$} (add2);
  \draw[flow] (10.15,0.6) -- (add2.south);
  \node[lbl, anchor=west] at (10.2,0.95) {$\Dcen$};

  \node[op] (add3) at (13.2,2.1) {};
  \draw[line width=0.6pt] ($(add3)+(-0.1,0)$) -- ($(add3)+(0.1,0)$)
                          ($(add3)+(0,-0.1)$) -- ($(add3)+(0,0.1)$);
  \draw[flow] (add2) -- (add3);
  \node[lbl, above] at (10.95,2.1) {$\Ahat$};
  \node[lbl, below] at (10.95,2.1) {\footnotesize(Eq.~\KRAeq{eq:residual_to_raw_inference})};

  \node[blk] (inn) at (13.95,3.85)
     {Innovation\\[-1pt] $\mathbf e(t)$ {\footnotesize(Eqs.~\KRAeq{eq:velocity_innovation}--\KRAeq{eq:innovation_correction})}};
  \node[lbl] (vo) at (12.3,3.85) {$\Vprof$};
  \draw[flow] (vo) -- (inn);
  \draw[flow] (inn.south -| add3) -- node[lbl, right] {$\mathbf c(t)$} (add3);

  \node[blk, minimum width=1.25cm] (rob) at (16.35,2.1) {Robot};
  \draw[flow] (add3) -- (rob);
  \node[lbl, above] at (14.55,2.1) {$\hat{\mathbf p}_h{+}\mathbf c(t)$};
  \node[lbl, below] at (14.55,2.1) {\footnotesize(Eq.~\KRAeq{eq:innovation_action})};
  \draw[flow] (rob.north) -- (rob.north |- inn.east) -- (inn.east);
  \node[lbl, above] at (15.75,3.85) {$\mathbf v(t)$};

  \draw[flow, draw=black!55] (16.35,5.2) -- (16.35,5.55) -- (8.85,5.55) -- (8.85,5.2);
  \node[fill=white, text=black!60, inner xsep=4pt, inner ysep=1pt] at (12.6,5.55)
     {\footnotesize next chunk};
\end{tikzpicture}
}
\caption{\textbf{Overview of KRA.} A shared motion projection converts the
current object velocity and the task-specific acceleration estimate at source time $t_0$ into a centered
physical displacement $\Dcen$; its normalized counterpart is
$\Dnorm=\Dcen\oslash\mathbf S$. \textbf{(1)} Fine-tuning subtracts $\Dcen$ from
each demonstrated chunk $\Achunk$ and trains the flow-based VLA $\vla$ on the
normalized residual chunk $\mathcal N(\Ares)$ with its standard flow-matching loss. \textbf{(2)} At
inference, the fine-tuned $\vla$ generates a residual chunk with $K$ Euler
steps, inverts normalization, and adds back $\Dcen$. \textbf{(3)} During execution, a
bounded correction $\mathbf c(t)$, computed from the deviation of the
observed velocity $\mathbf v(t)$ from the forecast $\Vprof$, adjusts the
commanded translation with a constant gate and bounded stepwise changes;
the correction restarts when a chunk becomes active. Neural
conditioning $\mathbf z$ includes current velocity as additional state; the forecast profile is an
input to the analytic branch.}
\label{fig:kra_pipeline}
\end{figure*}
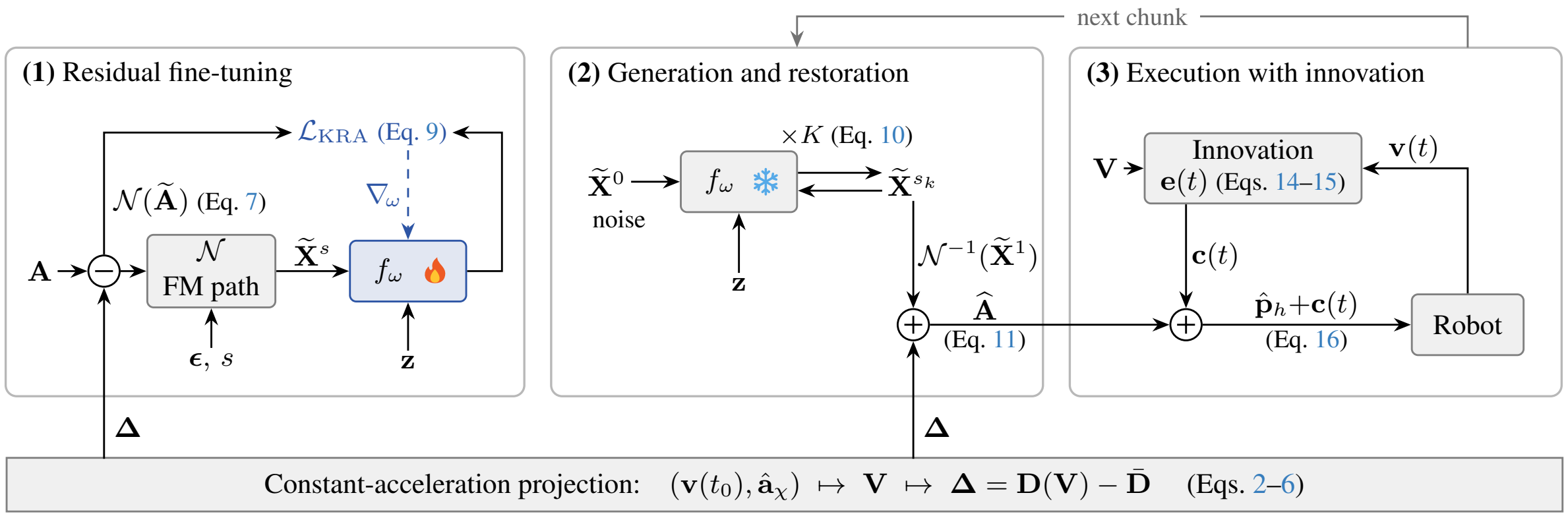

\subsection{Object Motion Projection}
\label{sec:prelim}
\label{subsec:problem_setting}
\label{subsec:kinematic_projection}

\noindent \textbf{Setup and notation.}
At source-observation time $t_0$, the policy receives an observation $\obs$ (RGB
images, a language instruction, and the proprioceptive state) and outputs a
chunk of $H$ future actions
$\Achunk=[\mathbf a_1,\ldots,\mathbf a_H]^\top\in\mathbb R^{H\times d_a}$.
Action $\mathbf a_h=(\mathbf p_h,\mathbf r_h,g_h)$ is scheduled for
execution at time $t_h$ and contains an end-effector translation target
$\mathbf p_h\in\mathbb R^3$, a quaternion orientation target $\mathbf r_h \in \mathbb{R}^4$, and a
gripper command $g_h$. We denote $\delta_j=t_{j+1}-t_j$ as the control interval, with
$\tau_h=t_h-t_0$ defined as scheduled action offset.
We consider translation targets expressed in one frame for the whole chunk,
either as absolute positions or relative to a common reference robot pose.

The target object moves with translational velocity
$\mathbf v(t)\in\mathbb R^3$, of which only the history
$\mathbf v_{\le t_0}$ is available at planning time. A demonstration dataset
$\mathcal D$ provides action chunks with observations and object-motion
records. Let $\chi$ identify the task's object--surface combination.
Throughout, $t$ denotes physical time and $s\in[0,1]$ the flow time of the
generative model. Planning uses only information available at $t_0$;
inference and chunk activation may occur later.

\vspace{3pt}
\noindent \textbf{Velocity-profile displacement (VPD).}
VPD is the planning-time forecast of how the object velocity evolves over
the chunk horizon. KRA needs it because the displacement assigned to each
action depends on object motion after $t_0$, which is not yet observed; a
constant-velocity profile ignores predictable speed changes such as
surface-induced deceleration.
Let
$\hat{\mathbf a}_0=\hat{\mathbf a}_{\chi}$ be the task-specific
object-acceleration estimate obtained from the dataset's motion records for
the corresponding object and surface. For real objects, this is a constant
deceleration profile estimated from repeated Monte Carlo runs on the
physical setup. The estimate is established before policy execution and
held fixed for the chunk. For elapsed times $\tau\ge0$,
\begin{equation}
 \hat{\mathbf v}_{t_0}(t_0+\tau)
 =\mathbf v(t_0)+\hat{\mathbf a}_0\tau.
 \label{eq:continuous_velocity_forecast}
\end{equation}
With $\tau_{H+1}$ denoting the end of the final control interval,
its action-aligned interval means are
\begin{align}
 \hat{\mathbf v}_j&=\mathbf v(t_0)
 +\tfrac12\hat{\mathbf a}_0(\tau_j+\tau_{j+1}),\notag\\
 \Vprof&=[\hat{\mathbf v}_1,\ldots,\hat{\mathbf v}_H]^\top
 =\psi(\mathbf v(t_0);\hat{\mathbf a}_{\chi})\in\mathbb R^{H\times3}.
 \label{eq:velocity_forecast}
\end{align}
Setting $\hat{\mathbf a}_0=\mathbf0$ gives the constant-velocity ablation.
The forecast is computed once at $t_0$ and held fixed while the chunk
executes. Velocity measurements that arrive later do not update it; they
are used only by the execution-time correction of
Sec.~\ref{subsec:velocity_innovation}.
The same task-specific acceleration estimate and forecast rule are used
with recorded current velocity to construct training targets and with live
current velocity at deployment. The current velocity is read
from simulator state in simulation and estimated by a causal visual tracker such as EfficientTAM~\cite{xiong2025efficient}
on the physical robot (Sec.~\ref{sec:exp_setup}).

\vspace{3pt}
\noindent \textbf{From velocity to action displacement.}
We define the forecast displacement of the object since the source
observation as
\begin{equation}
 \mathbf F(\tau)=\mathbf v(t_0)\tau
 +\tfrac12\hat{\mathbf a}_0\tau^2,\qquad \tau\ge0.
 \label{eq:forecast_displacement}
\end{equation}
Let $\tau_{\rm ref}$ be the offset from which the analytic action
projection is accumulated. The translation displacement for action $h$ is
\begin{equation}
 \mathbf d_h=\mathbf R\big[\mathbf F(\tau_h)
                      -\mathbf F(\tau_{\rm ref})\big].
 \label{eq:kinematic_projection}
\end{equation}
Here $\mathbf R\in SO(3)$ rotates measured velocity into the robot-base
frame of the translation targets. For DynamicVLA we accumulate displacement
from the first scheduled action, $\tau_{\rm ref}=\tau_1$, so that
$\mathbf d_h=\mathbf R\sum_{j=1}^{h-1}\delta_j\hat{\mathbf v}_j$. We call
this projection \emph{zero-based}: the first action receives no analytic
displacement ($\mathbf d_1=\mathbf0$), and object motion between the
observation and the first action is left to the learned residual.
Each adapter uses the same offsets and reference convention in
training and restoration.
Embedding the per-step displacements into the translational channels yields
the \emph{displacement chunk}
\begin{equation}
 \Dchunk(\Vprof)=[\mathbf d_1,\ldots,\mathbf d_H]^\top\mathbf P
 \in\mathbb R^{H\times d_a},
 \label{eq:displacement_chunk}
\end{equation}
where $\mathbf P=[\,\mathbf I_3\;\;\mathbf 0\,]\in\mathbb R^{3\times d_a}$
selects the translational channels, so the orientation and gripper channels
of $\Dchunk$ are zero.

\subsection{Residual Fine-Tuning}
\label{subsec:residual_target}

\noindent \textbf{Residual action chunks.}
KRA removes the projected displacement from every demonstrated chunk:
\begin{align}
 \Dcen&=\Dchunk(\Vprof)-\Dmean,
 \label{eq:centered_displacement}\\
 \Ares&=\Achunk-\Dcen,
 \label{eq:residual_target}
\end{align}
where the centering reference
$\Dmean=\frac{1}{M}\sum_{i=1}^{M}\Dchunk_i$ is the empirical mean of the
projected displacement chunks of the $M$ training chunks, held fixed during
validation and execution.
The \emph{residual chunk} $\Ares$ keeps the demonstrated orientation and
gripper channels. Its translation
$\tilde{\mathbf p}_h=\mathbf p_h-(\mathbf d_h-\bar{\mathbf d}_h)$, with
$\bar{\mathbf d}_h$ the corresponding block of $\Dmean$, is the demonstrated
target expressed in a frame that translates with the forecast object
motion. This removes the specified forecast displacement; residual approach,
contact timing, and forecast error can still depend on velocity. Centering by $\Dmean$ has no effect on the restored actions,
because the same $\Dcen$ is added back at inference. It preserves the mean
of the targets,
$\mathbb E_{\mathcal D}[\Ares]=\mathbb E_{\mathcal D}[\Achunk]$, without preserving their covariance or the full pretrained action
distribution.

The backbone is trained and sampled in normalized action space, so the
displacement must be expressed in the same units. With the backbone's fixed
affine action normalization $\mathcal N(\mathbf A)=(\mathbf A-\boldsymbol\mu)
\oslash\mathbf S$, where $\oslash$ denotes channelwise division, we define
$\widetilde{\mathbf X}=\mathcal N(\Ares)$ and $\mathbf X=\mathcal N(\Achunk)$.
Because $\mathcal N$ is affine, subtracting $\Dcen$ in physical units is
the same as subtracting its rescaled version $\Dnorm=\Dcen\oslash\mathbf S$
from the normalized chunk:
$\widetilde{\mathbf X}=\mathbf X-\Dnorm$. Thus $\Dnorm$ is the shift between
the raw and residual targets as seen by the network, and it is the quantity
used in the analysis of Sec.~\ref{subsec:residual_distribution}.

\vspace{3pt}
\noindent \textbf{Flow matching on residual chunks.}
Let $\vla$ denote any conditional flow matching-based VLA with parameters $\omega$, initialized
from pretrained weights. Given a noisy action chunk, the flow time, and the
conditioning, $\vla$ predicts the vector field that transports Gaussian
noise to action
chunks~\citep{lipman2023flow,tong2024improving}. KRA keeps this
model and its objective and changes only the data endpoint of the
probability path. With noise $\noise\sim\mathcal N(\mathbf 0,\mathbf I)$ and
flow time $s\sim p(s)$,
\begin{align}
 \widetilde{\mathbf X}^{s}&=(1-s)\,\noise+s\,\widetilde{\mathbf X},
 \label{eq:residual_path}\\
 \mathcal L_{\mathrm{KRA}}(\omega)&=
 \mathbb E\Big[\big\|\vla(\widetilde{\mathbf X}^{s},s,\mathbf z)
 -(\widetilde{\mathbf X}-\noise)\big\|_F^2\Big],
 \label{eq:krfm_loss}
\end{align}
where the expectation is over $\mathcal D$, $\noise$, and $s$, with
$p(s)=\mathcal U[0,1]$ the uniform flow-time sampling distribution and
$\|\cdot\|_F$ the Frobenius norm.
Here $\mathbf z$ denotes the conditioning fed to the network. It differs
from the observation $\obs$ only by the velocity input:
$\mathbf z=(\obs,\mathbf v(t_0))$ for velocity-conditioned models and
$\mathbf z=\obs$ otherwise.
By contrast, appending velocity to the
conditioning of a raw-action model changes neither the path nor the target.
In our primary experiments, each tested VLA backbone receives the current
three-dimensional object velocity $\mathbf v(t_0)$ as an additional input,
standardized using training-set velocity statistics and concatenated with
proprioception before the state-input projection. Thus
$\mathbf z$ contains the current velocity, not a future velocity profile.
The analytic branch uses $\Vprof$ independently of this neural input.
Velocity conditioning lets the residual model learn motion-dependent
approach and contact behavior and is ablated in Sec.~\ref{sec:exp_tracking}.

\subsection{Action Restoration}
\label{subsec:inference}

At planning time $t_0$, KRA computes $\Vprof$ and $\Dcen$ exactly as in
training and samples a residual chunk by integrating $\vla$ from
$\widetilde{\mathbf X}^{0}\sim\mathcal N(\mathbf 0,\mathbf I)$ with $K$ Euler steps at
$s_k=k/K$, $k=0,\ldots,K-1$:
\begin{equation}
 \widetilde{\mathbf X}^{s_{k+1}}=\widetilde{\mathbf X}^{s_k}
 +\tfrac{1}{K}\,\vla(\widetilde{\mathbf X}^{s_k},s_k,\mathbf z).
 \label{eq:residual_euler}
\end{equation}
The executable chunk is obtained by inverting
Eq.~\eqref{eq:residual_target}:
\begin{equation}
 \Ahat=\mathcal N^{-1}(\widetilde{\mathbf X}^{1})+\Dcen.
 \label{eq:residual_to_raw_inference}
\end{equation}
$\Dcen$ depends only on quantities available at $t_0$, so restoration is a
single addition and requires no additional network evaluation. Restoration
must be paired with residual training: adding $\Dcen$ to the output of a
policy trained on raw actions does not implement the inverse coordinate
transformation and can duplicate motion already represented in its output.

\subsection{Why Residual Coordinates?}
\label{subsec:residual_distribution}
\label{subsec:variance_reduction}
For a fixed projection, residualization is an invertible translation.
Let $\boldsymbol\zeta$ collect everything that defines the projection: the
observation, the current velocity, the task-specific acceleration estimate,
the action offsets and frame, and the centering reference. For a given
$\boldsymbol\zeta$, the shift $\Dnorm$ is therefore a known constant. Given a residual field
$\tilde f$ conditioned on $\boldsymbol\zeta$, define
\begin{equation}
 f(\mathbf X^s,s,\boldsymbol\zeta)
 =\tilde f(\mathbf X^s-s\Dnorm,s,\boldsymbol\zeta)+\Dnorm.
 \label{eq:flow_field_correspondence}
\end{equation}
Under shared noise, the paths obey
$\mathbf X^s=\widetilde{\mathbf X}^s+s\Dnorm$ and the regression errors
coincide. This establishes an exact correspondence between
unrestricted fields conditioned on the information defining the projection;
it does not guarantee approximation accuracy or improved optimization.

The learning motivation is to supply a predictable component analytically.
Write $\Achunk=\mathbf B+\Dchunk(\Vprof)$, where $\mathbf B$ is the remaining
behavior and may depend on velocity. At fixed context $\boldsymbol\kappa$ comprising the observation,
projection schedule, frame, and centering reference,
\begin{align}
 &\operatorname{Cov}(\Achunk\mid\boldsymbol\kappa)
 -\operatorname{Cov}(\Ares\mid\boldsymbol\kappa)\notag\\
 &\quad=\operatorname{Cov}(\Dchunk\mid\boldsymbol\kappa)+\mathbf C+\mathbf C^\top,
 \label{eq:covariance_difference}
\end{align}
where $\mathbf C=\operatorname{Cov}(\mathbf B,\Dchunk\mid\boldsymbol\kappa)$. Total variance decreases when the
right-hand side has positive trace; subtraction alone does not guarantee
this.

\subsection{Execution-time Velocity Innovation}
\label{subsec:velocity_innovation}
\begin{figure}[t]
\centering
\captionsetup{skip=3pt}
\begin{lrbox}{\paperfigurebox}\begin{minipage}{\columnwidth}
\centering
\includegraphics[width=0.85\linewidth]{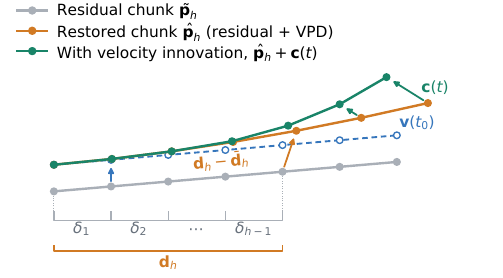}
\caption{Restoration and velocity innovation (one translation axis).
Gray: residual $\tilde{\mathbf p}_h$; dashed blue: constant-velocity
restoration under $\mathbf v(t_0)$; orange: restored $\hat{\mathbf p}_h$
using the forecast $\Vprof$; green: with the bounded execution-time offset
$\mathbf c(t)$. Brackets mark the control intervals $\delta_j$ and the
accumulated displacement $\mathbf d_h$, which is centered as
$\mathbf d_h-\bar{\mathbf d}_h$ before it is added to the residual.}
\label{fig:velocity_innovation}
\end{minipage}\end{lrbox}
\checkpaperfigure{velocity-innovation}
\end{figure}

A chunk generated at $t_0$ is reconstructed using the forecast $\Vprof$, which
stays frozen while the chunk is executed, whereas velocity measurements keep
arriving at the control rate. KRA uses them to correct the commanded
translation between replanning steps.
Figure~\ref{fig:velocity_innovation} illustrates this: the residual
translation $\tilde{\mathbf p}_h$ (gray) is restored with the planning-time
displacement $\mathbf d_h-\bar{\mathbf d}_h$ (orange), and the innovation
offset $\mathbf c(t)$ then moves the executed target toward the observed
object motion (green). With the continuous forecast of
Eq.~\eqref{eq:continuous_velocity_forecast}, defined from $t_0$ onward, the \emph{velocity innovation} (VI) at control
time $t$ is the accumulated difference between the observed and the forecast
motion in the action frame, i.e., the position error of the forecast:
\begin{equation}
 \mathbf e(t)=\mathbf R\int_{t_0}^{t}
 \big[\mathbf v(u)-\hat{\mathbf v}_{t_0}(u)\big]\,du .
 \label{eq:velocity_innovation}
\end{equation}
We use a constant gate of one: the analytic correction remains enabled
throughout execution, including gripper closing. To limit the magnitude
and stepwise change of the innovation offset, define
\begin{align}
 \mathbf c_k^*&=\sat_{r}(g\mathbf e(t_k)),\notag\\
 \mathbf c_k&=\mathbf c_{k-1}
  +\sat_{\eta}(\mathbf c_k^*-\mathbf c_{k-1}),
 \label{eq:innovation_correction}
\end{align}
where $\sat_m$ projects onto the Euclidean ball of radius $m$,
$g=1$ is a fixed gain, $r=0.05$\,m bounds the target offset, and
$\eta=0.005$\,m bounds its change per control step. The applied offset is
initialized at zero when a chunk becomes active.
The translation executed at time $t\in[t_h,t_{h+1})$ is
\begin{equation}
 \hat{\mathbf p}^{\rm exec}_h(t)=\hat{\mathbf p}_h+\mathbf c(t),
 \label{eq:innovation_action}
\end{equation}
where $\hat{\mathbf p}_h$ is the translation of the $h$-th action of
$\Ahat$; orientation and gripper commands are executed as generated. The
correction is an offset recomputed at every control step, not an increment
accumulated into the stored chunk, and its displacement-error anchor follows the source observation of the
active chunk. Integration and correction require no additional policy inference.

Innovation can
overcorrect systematic forecast errors already anticipated by the learned residual. Consequently, we treat it as a separately evaluated
component in Table~\ref{tab:innovation_ablation}.

\begin{figure*}[t]
  \centering
  \captionsetup{skip=3pt}
  \includegraphics[width=\textwidth]{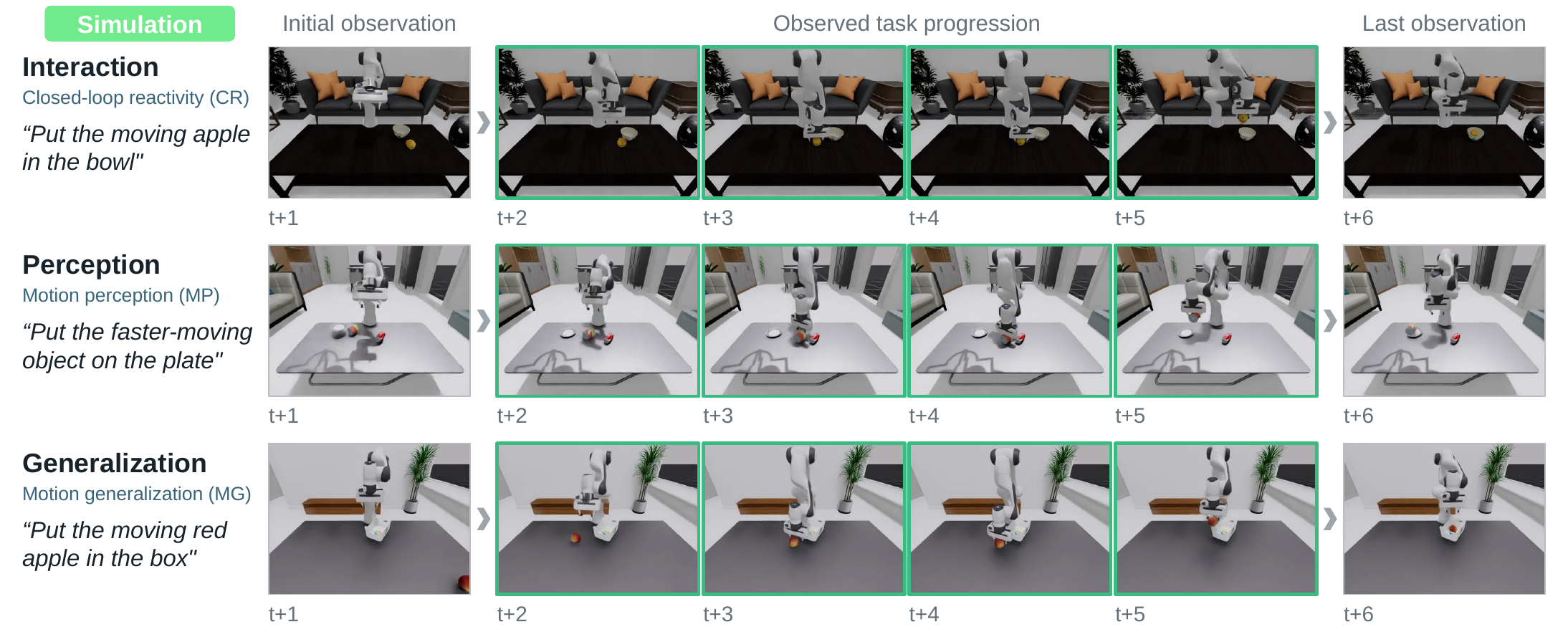}
  \caption{\textbf{Low-clutter DOM evaluation.}
  Selected successful \textbf{KRA} + DynamicVLA closed-loop rollouts
  in the low-clutter suite derived from DOM~\citep{xie2026dynamicvla}.
  Labels $t{+}1$--$t{+}6$ indicate successive snapshots rather than
  elapsed seconds.}
  \label{fig:demonstration_sequence}
\end{figure*}

\section{Experiments}
\label{sec:experiments}

We evaluate three questions: (1)~Does KRA improve moving-target tracking beyond neural velocity conditioning; (2)~Does it generalize to object speeds outside the training range; and (3)~Can it improve language-conditioned dynamic manipulation across VLA backbones and transfer to physical objects under causal visual velocity tracking?

\subsection{Benchmarks, Backbones, and Protocols}
\label{sec:exp_setup}

\noindent\textbf{Hover tracking.}
In NVIDIA Isaac Sim, a 7-DoF Franka Emika arm tracks a hover point 10\,cm above a moving apple. This controlled benchmark measures following accuracy before contact (\cref{sec:exp_tracking}).

\smallskip\noindent\textbf{Low-clutter DOM.}
Manipulation is evaluated on DOM~\cite{xie2026dynamicvla}, the dynamic-object manipulation dataset and benchmark, with scenes capped at five task objects. Its nine conditions cover Interaction: closed-loop reactivity (\textbf{CR}), dynamic adaptation (\textbf{DA}), and long-horizon sequencing (\textbf{LS}); Perception: visual understanding (\textbf{VU}), spatial reasoning (\textbf{SR}), and motion perception (\textbf{MP}); and Generalization: visual changes (\textbf{VG}), motion changes (\textbf{MG}), and disturbance robustness (\textbf{DR}). The simulator supplies the ground-truth velocity of the designated target object to every velocity-conditioned policy and to KRA's analytic branch.

\smallskip\noindent\textbf{Physical robot.}
A VX300-series ALOHA follower arm~\cite{zhao2023act} with calibrated scene and wrist RGB-D cameras intercepts an orange, a striped ball, or a green marker released from a ramp at approximately $0.3$ or $0.5$\,m/s and places it into a box. Current velocity is estimated causally from EfficientTAM~\cite{xiong2025efficient} masks lifted to 3D with calibrated depth.

\smallskip\noindent\textbf{Backbones and conditioning.}
We test \textbf{DynamicVLA}~\cite{xie2026dynamicvla} in all settings and extend low-clutter DOM to \textbf{SmolVLA}~\cite{shukor2025smolvla}, \textbf{VLASH}~\cite{tang2025vlash}, and \textbf{$\boldsymbol{\pi}_{0.5}$}~\cite{physicalintelligence2025pi05}. Unless marked ``no velocity'', every baseline and KRA model receives the current 3D object velocity as a neural input, and every \textbf{KRA} + backbone row on DOM and on the physical robot uses residual coordinates with VPD and VI (Sec.~\ref{sec:method}). Baselines keep their native action generation and streaming.

\begin{table*}[t]
\centering
\caption{Controlled hover tracking. Coverage is time within 3\,cm (\%); RMSE is mean per-episode root-mean-square position error (mm). Each condition has 200 matched trials per model; rows without neural velocity input are marked ``no velocity'', as in Table~\ref{tab:dom_main}.}
\label{tab:tracking}
\begingroup
\setlength{\tabcolsep}{4pt}
\renewcommand{\arraystretch}{1.08}
\footnotesize
\begin{tabular*}{\textwidth}{@{\extracolsep{\fill}}l*{8}{r}@{}}
\toprule
& \multicolumn{2}{c}{CR} & \multicolumn{2}{c}{DA} & \multicolumn{2}{c}{LS} & \multicolumn{2}{c}{Mean} \\
\cmidrule(lr){2-3}\cmidrule(lr){4-5}\cmidrule(lr){6-7}\cmidrule(l){8-9}
Model & Coverage $\uparrow$ & RMSE $\downarrow$ & Coverage $\uparrow$ & RMSE $\downarrow$ & Coverage $\uparrow$ & RMSE $\downarrow$ & Coverage $\uparrow$ & RMSE $\downarrow$ \\
\midrule
DynamicVLA (no velocity) & 10.94 & 66.25 & 10.40 & 68.10 & 8.84 & 68.16 & 10.06 & 67.50 \\
DynamicVLA & \textbf{53.66} & 34.87 & 45.73 & 38.20 & 46.81 & 36.62 & 48.74 & 36.57 \\
\textbf{KRA} + DynamicVLA (no velocity) & 51.38 & 34.93 & 50.52 & 36.73 & 49.71 & 34.45 & 50.54 & 35.37 \\
\textbf{KRA} + DynamicVLA & 52.87 & \textbf{33.41} & \textbf{55.18} & \textbf{34.16} & \textbf{52.10} & \textbf{33.17} & \textbf{53.38} & \textbf{33.58} \\
\bottomrule
\end{tabular*}
\endgroup
\end{table*}

\begin{table*}[!t]
\centering
\caption{Low-clutter DOM manipulation: task success (\%) and all-trial mean termination time (s). Each row evaluates a single checkpoint across 9 conditions $\times$ 200 trials = 1,800 trials. All primary baselines and their KRA counterparts are conditioned on current object velocity; KRA rows enable VPD and VI, and each backbone keeps its native asynchronous streaming. DynamicVLA rows are our fine-tuned checkpoints, not the published numbers.}
\label{tab:dom_main}
\begingroup
\setlength{\tabcolsep}{2.4pt}
\renewcommand{\arraystretch}{1.02}
\scriptsize
\begin{tabular*}{\textwidth}{@{\extracolsep{\fill}}l*{11}{r}@{}}
\toprule
& \multicolumn{3}{c}{Interaction} & \multicolumn{3}{c}{Perception}
& \multicolumn{3}{c}{Generalization} & \multicolumn{2}{c}{Average} \\
\cmidrule(lr){2-4}\cmidrule(lr){5-7}\cmidrule(lr){8-10}\cmidrule(l){11-12}
Method & CR & DA & LS & VU & SR & MP & VG & MG & DR & Success $\uparrow$ & Time $\downarrow$ \\
\midrule
DynamicVLA (no velocity) & 31.00 & 4.00 & \textbf{28.00} & 15.00 & 16.00 & 13.00 & 27.00 & 31.00 & 14.00 & 19.89 & 8.67 \\
DynamicVLA & 35.00 & 6.00 & 13.00 & 26.00 & 28.00 & 12.00 & 32.00 & 54.00 & 27.00 & 25.89 & 8.77 \\
\textbf{KRA} + DynamicVLA & \textbf{41.00} & \textbf{15.00} & 19.00 & \textbf{29.00} & 28.00 & \textbf{14.00} & \textbf{41.00} & \textbf{62.00} & \textbf{28.00} & \textbf{30.78} & 8.95 \\
\midrule
SmolVLA & 2.50 & 1.50 & 0.00 & 1.50 & 1.50 & 3.50 & 0.00 & 4.50 & 1.00 & 1.78 & 10.59 \\
\textbf{KRA} + SmolVLA & \textbf{3.00} & \textbf{8.50} & 0.00 & \textbf{4.50} & 1.50 & 3.50 & \textbf{3.50} & \textbf{6.00} & \textbf{1.50} & \textbf{3.56} & 10.56 \\
\midrule
VLASH & 7.00 & 5.00 & 1.00 & 7.00 & 4.00 & 7.00 & 3.00 & 9.00 & 1.00 & 4.89 & 10.00 \\
\textbf{KRA} + VLASH & \textbf{9.00} & \textbf{12.00} & 1.00 & \textbf{9.00} & \textbf{8.00} & 7.00 & \textbf{7.00} & \textbf{10.00} & 1.00 & \textbf{7.11} & 9.82 \\
\midrule
$\pi_{0.5}$ & 11.00 & 9.50 & 3.00 & 9.50 & 7.50 & 12.50 & 7.00 & 22.00 & 2.50 & 9.39 & 9.51 \\
\textbf{KRA} + $\pi_{0.5}$ & \textbf{15.00} & \textbf{30.00} & \textbf{21.50} & \textbf{24.00} & \textbf{15.00} & \textbf{19.00} & \textbf{8.50} & \textbf{28.00} & \textbf{10.50} & \textbf{19.06} & 9.57 \\

\bottomrule
\end{tabular*}
\endgroup
\end{table*}

\subsection{Controlled Hover Tracking Results}
\label{sec:exp_tracking}

\noindent\textbf{Protocol.}
Four DynamicVLA variants cross raw versus residual action coordinates with and without neural velocity input, sharing one training recipe on 300 demonstrations. Policies replan every $0.8$\,s with physics paused during inference, which isolates the coordinate parameterization from latency; VI and asynchronous streaming are disabled. Each variant runs 200 matched trials in CR (reversals, 10\,s), DA (varying 3-axis motion, 10\,s), and LS (30\,s sustained following), scored by \textbf{Coverage} (\% of time within 3\,cm) and 3D \textbf{RMSE} (mm).

\smallskip\noindent\textbf{Results.}
\Cref{tab:tracking} shows that, with shared neural velocity conditioning, residual coordinates raise mean coverage from $48.74\%$ to $53.38\%$ ($+4.64$\,pp) and reduce RMSE from $36.57$\,mm to $33.58$\,mm ($-8.2\%$). DynamicVLA holds slightly higher CR coverage ($53.66\%$ vs.\ $52.87\%$), while KRA + DynamicVLA has lower CR RMSE and leads on both metrics in DA ($55.18\%$ vs.\ $45.73\%$) and LS ($52.10\%$ vs.\ $46.81\%$). Without neural velocity inputs, KRA raises coverage from $10.06\%$ to $50.54\%$, although analytic restoration here also supplies velocity measurements that the unconditioned baseline never sees.

\smallskip\noindent\textbf{Speed generalization.}
In a separate sweep over 10 held-out paths (20 trials per speed, shared neural velocity inputs), KRA improves coverage $2.22\times$ at the training speed ($25.45\%$ to $56.46\%$) and $1.40\times$ across $1.5\times$ and $2\times$ out-of-distribution replay ($14.22\%$ to $19.96\%$), including $1.55\times$ at $2\times$ speed ($7.02\%$ to $10.89\%$). Gains are not uniform across metrics: at $2\times$ speed the raw-action baseline retains lower RMSE over the steps completed before termination.

\subsection{Low-Clutter DOM Manipulation}
\label{sec:exp_dom}

\noindent\textbf{Protocol.}
Each backbone is fine-tuned on the same 20,000-episode low-clutter DOM training subset. Each checkpoint is evaluated on ten configurations per condition with twenty trials each, i.e., 200 trials per condition and 1,800 per checkpoint, with continuous object motion and native asynchronous streaming. Success is the fraction of trials completing the instructed task, averaged over the nine conditions. Time is the mean termination time over all trials, not inference latency.

\smallskip\noindent\textbf{Results.}
In \Cref{tab:dom_main}, mean success increases from $25.89\%$ to $30.78\%$ for DynamicVLA, $1.78\%$ to $3.56\%$ for SmolVLA, $4.89\%$ to $7.11\%$ for VLASH, and $9.39\%$ to $19.06\%$ for $\pi_{0.5}$. DynamicVLA improves in eight conditions and ties in one; $\pi_{0.5}$ improves in all nine; SmolVLA and VLASH each improve in six and tie in three. For KRA + DynamicVLA, the largest absolute gains are in DA ($6\%$ to $15\%$), VG ($32\%$ to $41\%$), and MG ($54\%$ to $62\%$), yet DA, MP ($14\%$), and LS ($19\%$) remain difficult. Success counts do not separate target-selection errors from interception, grasping, or placement failures. Successful KRA + DynamicVLA rollouts are shown in Figure~\ref{fig:demonstration_sequence}.

\smallskip\noindent\textbf{Relation to published DynamicVLA results.}
The DynamicVLA rows are not the published numbers. We fine-tune the pretrained DynamicVLA checkpoint on our low-clutter training subset and test it on our low-clutter evaluation set, so its success ($25.89\%$) differs from the $47.06\%$ reported on unrestricted DOM~\cite{xie2026dynamicvla}. We did not isolate how much of this gap is due to the subset, the clutter cap, the compute budget, or hyperparameters, so the results measure adaptation within our protocol. A DynamicVLA checkpoint trained on the same split without neural velocity input reaches $19.89\%$; the comparison is not uniform across conditions, as this control scores higher on LS and MP.
\begin{table}[!t]
\centering
\caption{Real-robot CR pick-and-place completion at approximate ramp-exit speed $v$ (m/s) across 240 trials (20 per object $\times$ speed $\times$ method cell; Avg.\ pools 60 trials/method/speed). Entries are success (\%).}
\label{tab:real_robot_results}
\begingroup
\footnotesize
\setlength{\tabcolsep}{2.5pt}
\renewcommand{\arraystretch}{1.04}
\begin{tabular*}{\columnwidth}{@{\extracolsep{\fill}}lrrrrr@{}}
\toprule
Method & $v$ & Orange & Ball & Marker & Avg. \\
\midrule
DynamicVLA & 0.3 & 50.00 & 35.00 & 25.00 & 36.67 \\
           & 0.5 & 40.00 & 25.00 & 20.00 & 28.33 \\
\midrule
\textbf{KRA} + DynamicVLA & 0.3 & \textbf{65.00} & \textbf{40.00} & \textbf{35.00} & \textbf{46.67} \\
                          & 0.5 & \textbf{50.00} & \textbf{35.00} & \textbf{25.00} & \textbf{36.67} \\
\bottomrule
\end{tabular*}
\endgroup
\end{table}

\begin{figure*}[t]
  \centering
  \captionsetup{skip=3pt}
  \includegraphics[width=\textwidth]{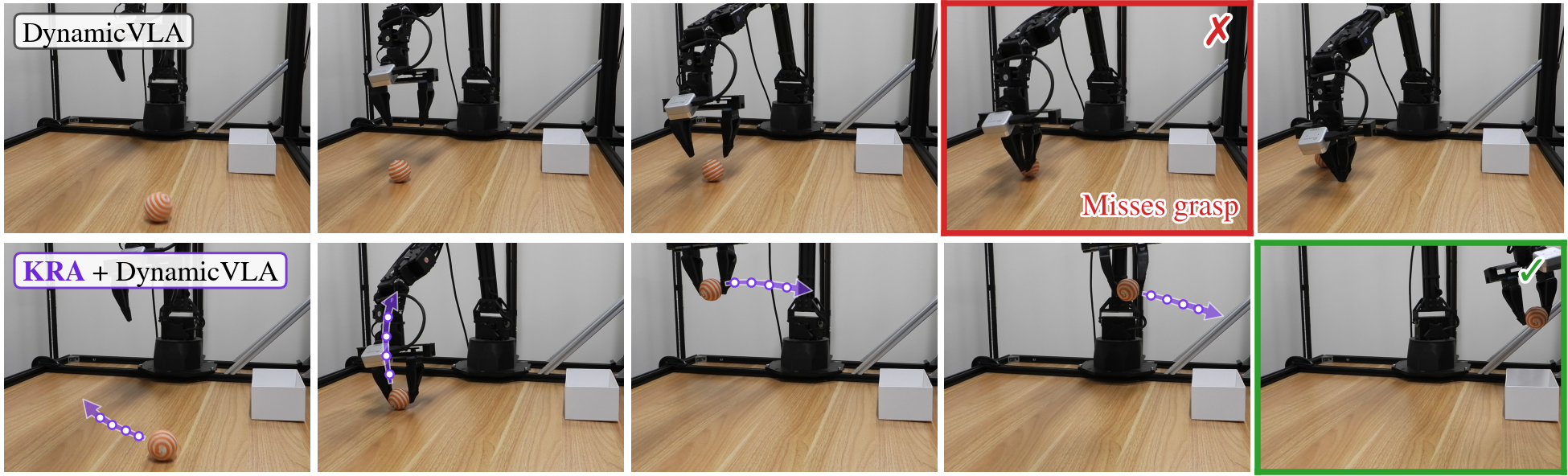}
  \caption{\textbf{Real-robot closed-loop evaluation.}
  Task progression for the same moving object with DynamicVLA (top), which
  misses the pickup, and \textbf{KRA} + DynamicVLA (bottom), which compensates
  the forecast object displacement and completes the pick.}
  \label{fig:real_policy_sequence}
\end{figure*}

\subsection{Real-Robot Pick-and-Place Evaluation}
\label{sec:exp_real}

\noindent\textbf{Protocol.}
Both policies are fine-tuned on the training partition of 900 real demonstrations (300 per object; 700/100/100 train/validation/test). We run twenty trials per object $\times$ speed $\times$ method cell, 240 trials in total, and report pick-and-place completion.

\smallskip\noindent\textbf{Results.}
\Cref{tab:real_robot_results} and Figure~\ref{fig:real_policy_sequence} summarize the physical CR trials with causal EfficientTAM velocity tracking. At $v\approx0.3$\,m/s, KRA + DynamicVLA raises pooled 60-trial success from $36.67\%$ to $46.67\%$ ($+10.00$\,pp); at $v\approx0.5$\,m/s, from $28.33\%$ to $36.67\%$ ($+8.33$\,pp). KRA has the higher completion count in every object and speed cell, although individual 20-trial cells carry substantial sampling variance. Pooling all six cells gives $39/120$ successes for the baseline versus $50/120$ for KRA ($32.50\%$ vs.\ $41.67\%$). A two-sided Fisher exact test under an independent-trial approximation yields $p=0.18$, a positive exploratory trend, not a statistically established gain.

\begin{table}[!t]
\centering
\caption{Component ablation on DynamicVLA over the nine low-clutter DOM conditions (1,800 trials per variant, shared neural velocity inputs). VPD: constant-acceleration forecast in place of the constant-velocity profile; VI: execution-time correction.}
\label{tab:innovation_ablation}
\begingroup
\small
\setlength{\tabcolsep}{4pt}
\renewcommand{\arraystretch}{1.00}
\begin{tabular*}{\columnwidth}{@{\extracolsep{\fill}}cccr@{}}
\toprule
KRA & VPD & VI & Success (\%) $\uparrow$ \\
\midrule
\textcolor{red!75!black}{$\times$} & \textcolor{red!75!black}{$\times$} & \textcolor{red!75!black}{$\times$} & 25.89 \\
\textcolor{green!45!black}{$\checkmark$} & \textcolor{red!75!black}{$\times$} & \textcolor{red!75!black}{$\times$} & 30.17 \\
\textcolor{green!45!black}{$\checkmark$} & \textcolor{green!45!black}{$\checkmark$} & \textcolor{red!75!black}{$\times$} & 30.56 \\
\textcolor{green!45!black}{$\checkmark$} & \textcolor{green!45!black}{$\checkmark$} & \textcolor{green!45!black}{$\checkmark$} & 30.78 \\
\bottomrule
\end{tabular*}
\endgroup
\end{table}

\begin{table}[!t]
\centering
\caption{Velocity-source sensitivity on low-clutter DOM over 1,800 trials per setting (success, \%). All columns use DynamicVLA; KRA uses VPD and VI, and KRA w/o VI retains VPD alone. Gaussian rows add per-axis noise $\sigma_v$ (m/s) to simulator velocity; the last row estimates velocity from EfficientTAM-Ti~\cite{xiong2025efficient} RGB-D masks. Bold: highest success per row.}
\label{tab:velocity_uncertainty}
\begingroup
\small
\setlength{\tabcolsep}{4pt}
\renewcommand{\arraystretch}{1.00}
\begin{tabular*}{\columnwidth}{@{\extracolsep{\fill}}lrrr@{}}
\toprule
Velocity source & DynamicVLA & KRA w/o VI & KRA \\
\midrule
$\sigma_v=0$ & 25.89 & 30.56 & \textbf{30.78} \\
$\sigma_v=0.01$ & 25.56 & 30.28 & \textbf{30.67} \\
$\sigma_v=0.05$ & \textbf{25.00} & 24.50 & 24.61 \\
$\sigma_v=0.10$ & \textbf{24.39} & 23.67 & 23.78 \\
\midrule
EfficientTAM-Ti & 26.11 & 29.33 & \textbf{31.00} \\
\bottomrule
\end{tabular*}
\endgroup
\end{table}

\subsection{Ablation Studies}
\label{sec:exp_ablations}

\noindent\textbf{VPD and VI.}
\Cref{tab:innovation_ablation} ablates the DynamicVLA components under the low-clutter DOM protocol above. The raw-action baseline attains $25.89\%$. KRA with a constant-velocity profile and no VI reaches $30.17\%$ ($+4.28$\,pp); the constant-acceleration forecast (VPD) reaches $30.56\%$ ($+0.39$\,pp); and adding VI reaches $30.78\%$ ($+0.22$\,pp). Two-sided Fisher exact tests under an independent-trial approximation give $p\approx0.0048$, $0.83$, and $0.91$ for successive rows and $p\approx0.0013$ for baseline versus full KRA. Only the residual-training step is statistically supported; the increments from VPD and VI are within sampling noise, so their individual benefits are not established.

\smallskip\noindent\textbf{Velocity source.}
\Cref{tab:velocity_uncertainty} first adds per-axis Gaussian noise with standard deviation $\sigma_v$ to the simulator velocity. At $\sigma_v=0.01$\,m/s, full KRA retains $30.67\%$ ($-0.11$\,pp) and exceeds DynamicVLA by $5.11$\,pp. At $\sigma_v=0.05$ and $0.10$\,m/s the ranking reverses: DynamicVLA reaches $25.00\%$ and $24.39\%$ versus $24.61\%$ and $23.78\%$ for full KRA. From clean input to $0.10$\,m/s noise, KRA loses $7.00$\,pp and DynamicVLA $1.50$\,pp, so the benefit of explicit motion compensation depends on velocity quality. With velocity estimated causally from EfficientTAM-Ti~\cite{xiong2025efficient} RGB-D masks in place of simulator velocity, KRA succeeds in $558$ of $1{,}800$ trials ($31.00\%$) versus $470$ ($26.11\%$) for DynamicVLA, close to the clean-velocity scores, which suggests that KRA does not require privileged simulator velocity. KRA without VI reaches $528$ ($29.33\%$), so VI contributes $1.67$\,pp here versus $0.22$\,pp with clean velocity.

\section{Limitations}
\label{sec:limitations}
VPD forecasts object motion from observations available at planning time;
forecast errors, contact dynamics, joint limits, and collisions can make the
required action shift differ from this analytic displacement. VI measures
departure from the planned profile, not error in the neural policy, so it
can overcorrect motion already anticipated by the residual, and its bounds
do not establish collision avoidance or continuity when it restarts at
chunk activation. KRA depends on velocity quality and falls below DynamicVLA
at the two larger noise levels.
Simulation supplies privileged target velocity, which limits conclusions
about unaided target identification. Tracking pauses simulation during
inference, so its gains do not by themselves establish robustness to
deployment latency. Results use single checkpoints and do not characterize
variation across training seeds. The physical study covers three objects
and two speed tiers, its success increase is exploratory, and
visual-tracker latency is unreported.

\section{Conclusion}
\label{sec:conclusion}
KRA combines a planning-time velocity-profile displacement with a learned
manipulation residual: one projection defines both the residual training
targets and action restoration, and a bounded velocity innovation corrects
departures from the forecast without another policy inference. Tracking
improves under shared velocity conditioning, including at held-out speeds,
mean manipulation success increases across four VLA backbones on low-clutter
DOM, and physical pick-and-place shows a positive exploratory trend.

\section*{LLM Usage Statement}
The authors take full responsibility for all content of this paper. Anthropic
Claude was the only LLM or generative AI tool used in this work, as follows.
\textbf{Writing and editing text:} Claude assisted by editing, restructuring,
and condensing author-written text.
\textbf{Related work and citations:} Claude generated draft related-work text
and citation entries, which the authors reviewed, edited, and approved.
\textbf{Code and implementation:} Claude generated code, which the authors
reviewed, edited, and approved.
\textbf{Experiments (design, execution, analysis of results):} Claude assisted
by refining author-produced analysis and checking the consistency of reported
numbers.
\textbf{Figures, diagrams, and visual material:} Claude assisted by refining
author-produced figures.
\textbf{Data (synthetic data, dataset creation, annotation):} no LLM or AI-tool
use.
\textbf{Research ideas, hypotheses, claims, and conclusions:} no LLM or AI-tool
use.
The manuscript was not given to an LLM provider to obtain reviewer-style
feedback.

{\small
\bibliographystyle{ieeenat_fullname}
\bibliography{main}
}

% ---------------------------------------------------------------------------
% Appendix: the supplementary material. Sections are lettered A-I as in the
% separate supplement; figures, tables, algorithms, theorems and equations
% carry the prefix S so they cannot be confused with the main paper.
% ---------------------------------------------------------------------------
\twocolumn[{\centering\Large\bfseries Supplementary Material\par\vspace{1.5em}}]
\appendix
\setcounter{equation}{0}
\setcounter{figure}{0}
\setcounter{table}{0}
\setcounter{algorithm}{0}
\setcounter{theorem}{0}
\renewcommand{\theequation}{S\arabic{equation}}
\renewcommand{\thefigure}{S\arabic{figure}}
\renewcommand{\thetable}{S\arabic{table}}
\renewcommand{\thealgorithm}{S\arabic{algorithm}}
\renewcommand{\thetheorem}{S\arabic{theorem}}
% Distinct hyperlink anchors for the restarted counters.
\def\theHequation{S\arabic{equation}}
\def\theHfigure{S\arabic{figure}}
\def\theHtable{S\arabic{table}}
\def\theHalgorithm{S\arabic{algorithm}}
\def\theHtheorem{S\arabic{theorem}}
\makeatletter\def\theHALG@line{\thealgorithm.\arabic{ALG@line}}\makeatother
\def\includetrackingdetails{}

\noindent This appendix supplements the main paper.
Sec.~\ref{supp:alg} gives the complete algorithms of KRA,
Sec.~\ref{supp:details} its implementation conventions together with the
measured ramp-release speed profiles of the physical objects, and
Sec.~\ref{supp:analysis} the derivations behind
Sec.~\mainref{subsec:residual_distribution}, including a theorem on the
covariance of residual targets.
Secs.~\ref{supp:ablation}, \ref{supp:tracking}, and~\ref{supp:exp_details}
give the protocols of the ablations, the hover-tracking study, and the
manipulation and real-robot experiments; Sec.~\ref{supp:runtime} states the
computational cost, Sec.~\ref{supp:limitations} further limitations, and
Sec.~\ref{supp:video} is a guide to the supplementary video.
Figure, table, algorithm, theorem, and equation numbers with the prefix
``S'' refer to this appendix; all others refer to the main paper.

\section{Detailed Algorithms}
\label{supp:alg}
This section gives the complete procedures of KRA described in
Sec.~\mainref{sec:method}: the shared motion projection
(Algorithm~\ref{alg:supp_project}), residual fine-tuning
(Algorithm~\ref{alg:supp_train}), chunk generation with action restoration
(Algorithm~\ref{alg:supp_infer}), and execution-time velocity innovation
(Algorithm~\ref{alg:supp_vi}). Equation numbers refer to the main paper, and
Table~\ref{tab:supp_notation} collects its notation.
Throughout, $\sat_m(\mathbf x)=\mathbf x\,\min(1,m/\|\mathbf x\|_2)$ is the
projection onto the Euclidean ball of radius $m$.

\begin{table}[t]
\centering
\caption{Notation of the main paper used in the algorithms.}
\label{tab:supp_notation}
\footnotesize
\setlength{\tabcolsep}{3pt}
\renewcommand{\arraystretch}{1.08}
\begin{tabular}{@{}p{0.31\columnwidth}p{0.66\columnwidth}@{}}
\toprule
Symbol & Meaning \\
\midrule
$t_0$; $\tau_h$; $\delta_j$ & source-observation time; offset $t_h-t_0$ of action $h$; control interval $t_{j+1}-t_j$ \\
$\obs$; $\mathbf z$ & observation; neural inputs of the backbone (images, language, proprioception with standardized current velocity) \\
$\Achunk\in\mathbb R^{H\times d_a}$ & demonstrated chunk, $\mathbf a_h=(\mathbf p_h,\mathbf r_h,g_h)$ \\
$\mathbf v(t)$; $\chi$; $\hat{\mathbf a}_\chi$ & object velocity; object--surface combination; its fixed acceleration estimate \\
$\Vprof=\psi(\mathbf v(t_0);\hat{\mathbf a}_\chi)$ & forecast velocity profile (interval means $\hat{\mathbf v}_j$) \\
$\mathbf F(\tau)$; $\tau_{\rm ref}$ & forecast displacement since $t_0$; offset from which the projection is accumulated \\
$\mathbf R$; $\mathbf P$ & rotation into the frame of the translation targets; selector of the translational channels \\
$\Dchunk(\Vprof)$; $\Dmean$; $\Dcen$ & displacement chunk; its training-set mean; centered displacement \\
$\Ares$; $\Ahat$ & residual chunk; restored (executable) chunk \\
$\mathcal N$; $\boldsymbol\mu$, $\mathbf S$ & fixed affine action normalization of the backbone and its statistics \\
$\mathbf X$; $\widetilde{\mathbf X}$; $\Dnorm$ & $\mathcal N(\Achunk)$; $\mathcal N(\Ares)$; $\Dcen\oslash\mathbf S$ \\
$\vla$; $\noise$; $s$; $K$ & flow-based VLA; base noise; flow time; Euler steps \\
$\mathbf e(t)$; $\mathbf c_k$ & velocity innovation; applied translation offset \\
$g$; $r$; $\eta$ & gain ($1$); offset bound ($0.05$\,m); per-step bound ($0.005$\,m) \\
\bottomrule
\end{tabular}
\end{table}

\begin{algorithm}[t]
\caption{Motion projection (shared by training and deployment)}
\label{alg:supp_project}
\small
\begin{algorithmic}[1]
\Function{Project}{$\mathbf v(t_0),\,\hat{\mathbf a}_0,\,\tau_{1:H},\,\tau_{\rm ref}$}
  \For{$h=1,\ldots,H$}
    \State $\mathbf F_h\gets\mathbf v(t_0)\tau_h+\tfrac12\hat{\mathbf a}_0\tau_h^2$
      \Comment{\meq{eq:forecast_displacement}}
  \EndFor
  \State $\mathbf F_{\rm ref}\gets\mathbf v(t_0)\tau_{\rm ref}
     +\tfrac12\hat{\mathbf a}_0\tau_{\rm ref}^2$
  \For{$h=1,\ldots,H$}
    \State $\mathbf d_h\gets\mathbf R\,(\mathbf F_h-\mathbf F_{\rm ref})$
      \Comment{\meq{eq:kinematic_projection}}
  \EndFor
  \State \Return $\Dchunk\gets[\mathbf d_1,\ldots,\mathbf d_H]^\top\mathbf P$
      \Comment{\meq{eq:displacement_chunk}}
\EndFunction
\end{algorithmic}
\end{algorithm}

\paragraph{Projection.}
Because $\mathbf F(\tau_{j+1})-\mathbf F(\tau_j)=\delta_j\hat{\mathbf v}_j$
for the interval means of \meq{eq:velocity_forecast}, the zero-based
convention $\tau_{\rm ref}=\tau_1$ gives
$\mathbf d_h=\mathbf R\sum_{j=1}^{h-1}\delta_j\hat{\mathbf v}_j$; setting
$\hat{\mathbf a}_0=\mathbf 0$ gives the constant-velocity variant of
Table~\mainref{tab:innovation_ablation}.

\subsection{Offline preparation and residual fine-tuning}
\label{supp:alg_train}
Algorithm~\ref{alg:supp_train} first computes the centering reference from
the training chunks and then fine-tunes the pretrained backbone on residual
chunks. The projection is applied to the \emph{recorded} current velocity of
each training chunk with the same acceleration estimate, offsets, and
reference convention as at deployment; future ground-truth velocities are
never supplied. The reference $\Dmean$ is one $H\times d_a$ tensor per
training recipe and, like the raw-action normalization statistics
$(\boldsymbol\mu,\mathbf S)$ and the velocity statistics of the neural
input, is fixed after this step.
Compared with raw-action fine-tuning, only
the two lines that compute $\Dcen$ and $\Ares$ are added.

\begin{algorithm}[t]
\caption{Residual fine-tuning for the objective in \meq{eq:krfm_loss}}
\label{alg:supp_train}
\small
\begin{algorithmic}[1]
\Require Pretrained flow-based VLA $\vla$; training chunks
 $\{(\obs_i,\mathbf v_i(t_0),\chi_i,\Achunk_i)\}_{i=1}^{M}$ with action
 offsets $\tau_{1:H}$; task estimates $\{\hat{\mathbf a}_\chi\}$;
 normalization $\mathcal N=(\boldsymbol\mu,\mathbf S)$
\Statex \textbf{Offline preparation}
\For{$i=1,\ldots,M$}
  \State $\Dchunk_i\gets\textsc{Project}(\mathbf v_i(t_0),
     \hat{\mathbf a}_{\chi_i},\tau_{1:H},\tau_{\rm ref})$
\EndFor
\State $\Dmean\gets\frac1M\sum_{i=1}^{M}\Dchunk_i$;\;
   store $\Dmean$ with the checkpoint
\Statex \textbf{Fine-tuning} (repeat until convergence)
\State Sample $(\obs,\mathbf v(t_0),\chi,\Achunk)$ from the training
   chunks, $\noise\sim\mathcal N(\mathbf 0,\mathbf I)$, $s\sim\mathcal U[0,1]$
\State $\Dcen\gets\textsc{Project}(\mathbf v(t_0),\hat{\mathbf a}_{\chi},
   \tau_{1:H},\tau_{\rm ref})-\Dmean$
   \Comment{\meq{eq:centered_displacement}}
\State $\Ares\gets\Achunk-\Dcen$;\;
   $\widetilde{\mathbf X}\gets\mathcal N(\Ares)$
   \Comment{\meq{eq:residual_target}}
\State $\widetilde{\mathbf X}^{s}\gets(1-s)\,\noise+s\,\widetilde{\mathbf X}$
   \Comment{\meq{eq:residual_path}}
\State $\mathbf z\gets$ neural inputs: $\obs$ with the standardized
   current velocity $\mathbf v(t_0)$ appended to the state
\State Update $\omega$ with
   $\nabla_{\omega}\big\|\vla(\widetilde{\mathbf X}^{s},s,\mathbf z)
   -(\widetilde{\mathbf X}-\noise)\big\|_F^2$
   \Comment{\meq{eq:krfm_loss}}
\Ensure Fine-tuned $\vla$; fixed $\Dmean$, $(\boldsymbol\mu,\mathbf S)$
\end{algorithmic}
\end{algorithm}

\subsection{Chunk generation and action restoration}
\label{supp:alg_infer}
Algorithm~\ref{alg:supp_infer} runs whenever the backbone plans a new chunk,
with the same projection as in training applied to the live current
velocity. The source time and the planned profile are stored with the chunk
because the velocity innovation needs them later, and the backbone keeps its
native scheduling of chunks (queueing, streaming, and replacement).

\begin{algorithm}[t]
\caption{Chunk generation and action restoration}
\label{alg:supp_infer}
\small
\begin{algorithmic}[1]
\Require Fine-tuned $\vla$; $\Dmean$; $\mathcal N=(\boldsymbol\mu,\mathbf S)$;
 task estimate $\hat{\mathbf a}_\chi$; Euler steps $K$
\Statex At each planning time $t_0$:
\State Read the observation $\obs$ and the current velocity $\mathbf v(t_0)$;
   form $\mathbf z$
\State $\Dcen\gets\textsc{Project}(\mathbf v(t_0),\hat{\mathbf a}_{\chi},
   \tau_{1:H},\tau_{\rm ref})-\Dmean$
   \Comment{as in training}
\State $\widetilde{\mathbf X}^{0}\sim\mathcal N(\mathbf 0,\mathbf I)$
\For{$k=0,\ldots,K-1$} \Comment{\meq{eq:residual_euler}}
  \State $\widetilde{\mathbf X}^{(k+1)/K}\gets\widetilde{\mathbf X}^{k/K}
    +\tfrac1K\,\vla(\widetilde{\mathbf X}^{k/K},k/K,\mathbf z)$
\EndFor
\State $\Ahat\gets\mathcal N^{-1}(\widetilde{\mathbf X}^{1})+\Dcen$
   \Comment{\meq{eq:residual_to_raw_inference}}
\State Attach to the chunk: source time $t_0$, the planned profile
   $(\mathbf v(t_0),\hat{\mathbf a}_\chi)$, and the offsets $\tau_{1:H}$
\State Pass $\Ahat$ to the backbone's native chunk scheduler
\Ensure Executable chunk $\Ahat$ with its source time and planned profile
\end{algorithmic}
\end{algorithm}

\subsection{Execution-time velocity innovation}
\label{supp:alg_vi}
Algorithm~\ref{alg:supp_vi} runs at the control rate between policy
inferences. It integrates the observed velocity once into a prefix integral
$\boldsymbol\Phi$, so that the innovation of \meq{eq:velocity_innovation}
for any active chunk is a difference of two prefix values minus the
forecast displacement.

\begin{algorithm}[t]
\caption{Execution-time velocity innovation (VI)}
\label{alg:supp_vi}
\small
\begin{algorithmic}[1]
\Require Gain $g=1$; offset bound $r=0.05$\,m; per-step bound
 $\eta=0.005$\,m; rotation $\mathbf R$
\Statex \textbf{State:} prefix integral $\boldsymbol\Phi$ of the observed
 velocity; applied offset $\mathbf c$; active chunk $\Ahat$ with source time
 $t_0$ and planned profile $(\mathbf v(t_0),\hat{\mathbf a}_0)$
\Statex \textbf{For every timestamped velocity sample} $(t_k,\mathbf v(t_k))$
   (trapezoidal rule):
\State $\boldsymbol\Phi(t_k)\gets\boldsymbol\Phi(t_{k-1})
   +\tfrac12\big(\mathbf v(t_{k-1})+\mathbf v(t_k)\big)(t_k-t_{k-1})$
\Statex \textbf{When a chunk becomes active:}
\State $\mathbf c\gets\mathbf 0$;\; load its $t_0$ and planned profile
   \Comment{reset at activation}
\Statex \textbf{At every control step} $t_k\in[t_h,t_{h+1})$:
\State $\mathbf F\gets\mathbf v(t_0)(t_k-t_0)
   +\tfrac12\hat{\mathbf a}_0(t_k-t_0)^2$
   \Comment{\meq{eq:forecast_displacement}}
\State $\mathbf e\gets\mathbf R\big[\boldsymbol\Phi(t_k)
   -\boldsymbol\Phi(t_0)-\mathbf F\big]$
   \Comment{\meq{eq:velocity_innovation}}
\State $\mathbf c^*\gets\sat_r(g\,\mathbf e)$;\;
   $\mathbf c\gets\mathbf c+\sat_\eta(\mathbf c^*-\mathbf c)$
   \Comment{\meq{eq:innovation_correction}}
\State Execute $\hat{\mathbf a}_h$ with translation
   $\hat{\mathbf p}_h+\mathbf c$; orientation and gripper as generated
   \Comment{\meq{eq:innovation_action}}
\end{algorithmic}
\end{algorithm}

\paragraph{Anchor.}
The innovation anchor remains $t_0$ even when $\tau_{\rm ref}>0$: VI
measures the error of the source-time motion forecast, including the
interval spent on inference, whereas the zero-based projection leaves the
motion before the first action to the learned residual. The planned profile
and the observed integral must therefore use the same source anchor.

\paragraph{Activation and streaming.}
The reset of $\mathbf c$ is tied to chunk activation, not to the completion
of inference for a queued chunk. Under streaming, a newly generated chunk
need not yet be active; actions skipped as stale keep their original
timestamps, so the observed-motion integral includes the elapsed inference
time rather than silently discarding it. The per-step limit $\eta$ applies
within an active chunk; the reset at a chunk boundary is a separate
operation and does not guarantee continuity of the executed trajectory.
Because the offset follows $\mathbf c^*$ under the rate limit, a previously
applied offset can persist briefly after the forecast error returns to zero.

\section{Method Details}
\label{supp:details}
This section gives the implementation conventions behind
Sec.~\mainref{sec:method}: action frames and timing, the velocity sources and
the forecast, centering and normalization, and neural velocity conditioning.

\subsection{Action conventions, frames, and timing}
\label{supp:conventions}
\paragraph{Action representation.}
The DOM adapters use the first three action channels for Cartesian
translation relative to a common reference pose, not successive per-step
increments; orientation and gripper channels keep the backbone's
representation. For VLASH, the reference can advance to the expected
execution state, so its offsets $\tau_h$ include the corresponding
observation-to-reference delay.

\paragraph{Reference offset.}
The DynamicVLA projection is zero-based ($\tau_{\rm ref}=\tau_1$,
$\mathbf d_1=\mathbf 0$). In hover tracking the action labels begin at
$\tau_1=0.2$\,s, and the displacement before that first action stays in the
learned residual. A projection that also removes the observation-to-action
displacement would instead use $\tau_{\rm ref}=0$. The forecast itself is
defined from the source observation onward under both conventions, which is
why VI (Sec.~\ref{supp:alg_vi}) is anchored at $t_0$. The scheduled offsets
remain attached to the chunk when asynchronous streaming skips stale
actions.

\paragraph{Frames.}
Simulator-world velocities are rotated into the robot base with the
world-to-base rotation, and camera-frame measurements with the calibrated
extrinsics at their measurement timestamps. After this conversion every
integrated velocity is expressed in the robot-base frame of the translation
targets, so $\mathbf R=\mathbf I_3$ in the implementation. The analytic
branch uses a constant gate of one in both training and execution.

\subsection{Velocity sources and forecast}
\label{supp:forecast}
\paragraph{Information available to the forecast.}
The forecast uses only the current velocity and the fixed task estimate
$\hat{\mathbf a}_\chi$; future ground-truth or measured velocities are never
supplied to $\psi$. Using the same construction rule for training targets
and deployment avoids a difference in available information between the
two; it does not imply identical forecast-error distributions.

\paragraph{Simulation.}
The current velocity of the designated target object is read from the
simulator state, and $\hat{\mathbf a}_\chi$ is estimated from the dataset's
motion records for each object--surface combination.

\paragraph{Physical robot.}
EfficientTAM~\citep{xiong2025efficient} segments the object in the RGB-D
stream of the calibrated scene camera. For the scored trials, we initialize
EfficientTAM with a previously acquired mask of the target object in view on
the ramp. Valid masked depth is back-projected with the camera intrinsics,
transformed to the robot base, and processed by a timestamped
position--velocity filter. When the tracker loses the mask, we forecast the
object's position and velocity with the calibrated deceleration for that
object and ramp height.

Objects are released from rest on inclined ramps of different heights to set
the nominal ramp-exit speed. To calibrate the forecast, we use repeated
grasping runs for each object and ramp height. In each run, we use the
interval from peak tracked speed to grasp and fit slopes to the three
robot-base velocity components. The repeated runs give one fixed
deceleration vector per object and ramp height. KRA uses the same calibrated
vector in its planning-time forecast.

\begin{figure}[t]
  \centering
  \includegraphics[width=\linewidth]{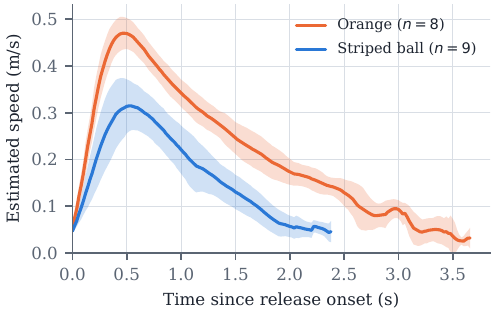}
  \caption{\textbf{Online EfficientTAM speed estimates for two ramp conditions.}
  The scene RGB-D tracker reports the magnitude of filtered 3D velocity for
  the orange in the $v\approx0.5$\,m/s condition and the striped ball in the
  $v\approx0.3$\,m/s condition. Runs are aligned at detected motion onset.
  Lines show means and bands show pointwise 95\% Student-$t$ confidence intervals
  for those means. Of ten recorded releases per object, we exclude runs whose
  peak speed or time to peak has an absolute modified $z$-score above 3.5
  (two orange runs and one ball run in which tracking stopped after 0.5\,s).
  Curves end once fewer than half of the retained runs have samples.}
  \label{fig:supp_ramp_velocity}
\end{figure}

\paragraph{Ramp speed profiles.}
For these measurements, an operator draws a box around the target in the
front camera to prepare a saved reference for EfficientTAM. A speed sample
requires an accepted mask and valid masked depth. For the plot, we resample
valid estimates on a 20\,ms grid and interpolate across gaps of at most
0.10\,s. Longer gaps are left undefined.
Figure~\ref{fig:supp_ramp_velocity} illustrates the causal velocity signal
available to the policies on the robot. In both examples the estimated speed
rises on the ramp, peaks near the exit, and falls as the object rolls across
the table. The mean peak speeds are $0.47\pm0.03$\,m/s for the orange ($n=8$)
and $0.33\pm0.04$\,m/s for the striped ball ($n=9$), reached $0.47\pm0.03$\,s
and $0.51\pm0.10$\,s after detected motion onset (mean $\pm$ 95\% confidence
interval half-width). These traces illustrate that the tracker captures the
rise and subsequent decrease in speed online. The speeds are tracker estimates
without an independent ground-truth reference.

\subsection{Centering reference and normalization}
\label{supp:centering}
$\Dmean$ is computed once from the projected displacement chunks of all
$M$ training chunks of a recipe, each constructed from the chunk's recorded
motion, task estimate, and action offsets with the same projection as at
deployment. All acceleration and timing contributions are included before
averaging, and the result is one $H\times d_a$ tensor shared across the
recipe's tasks. Because it is computed from the training chunks, the mean
preservation
$\mathbb E_{\mathcal D}[\Ares]=\mathbb E_{\mathcal D}[\Achunk]$ holds over
those chunks; it is not asserted under a shifted test distribution. The
raw-action normalization statistics $(\boldsymbol\mu,\mathbf S)$ and $\Dmean$
are fixed at evaluation and are not recomputed from residual predictions or
test motion; the same stored tensors are used for subtraction and
restoration.

\subsection{Neural velocity conditioning}
\label{supp:conditioning}
In our velocity-conditioned variants, the current object velocity enters the
backbone as three additional state channels. We standardize it using training-set
velocity statistics and concatenate it with proprioception before the state-input
projection. The future profile $\Vprof$ feeds only the analytic branch.
KRA uses the fine-tuning recipe of the corresponding backbone
(Sec.~\ref{supp:exp_details}) and changes the
translational supervision. The task estimates $\hat{\mathbf a}_\chi$ are
fixed calibration quantities.

\section{Analysis of Residual Coordinates}
\label{supp:analysis}
This section derives the two statements of
Sec.~\mainref{subsec:residual_distribution}, the residual-flow
correspondence (\meq{eq:flow_field_correspondence}) and the covariance
identity (\meq{eq:covariance_difference}), and states their scope. The
analysis concerns the residual generator before execution-time innovation.

\subsection{Residual-flow correspondence}
\label{supp:analysis_flow}
Let $\boldsymbol\zeta$ collect the observation, current velocity,
task-specific acceleration estimate, projection timing, frame, and centering
reference, so that the normalized shift $\Dnorm=\Dcen\oslash\mathbf S$ is a
known constant given $\boldsymbol\zeta$, and let $f$ be defined from a
residual field $\tilde f$ by \meq{eq:flow_field_correspondence}.

\paragraph{Paths and regression errors.}
For shared base noise $\noise$ and flow time $s\in[0,1]$, the raw and
residual interpolation paths are
\begin{align}
 \mathbf X^s&=(1-s)\,\noise+s\,\mathbf X,\qquad
 \widetilde{\mathbf X}^s=(1-s)\,\noise+s\,\widetilde{\mathbf X},\notag\\
 \mathbf X^s&=\widetilde{\mathbf X}^s+s\Dnorm,
 \label{eq:supp_path_relation}
\end{align}
where the last line uses $\mathbf X=\widetilde{\mathbf X}+\Dnorm$. Since
$\mathbf X-\noise=\widetilde{\mathbf X}-\noise+\Dnorm$,
\begin{align}
 &f(\mathbf X^s,s,\boldsymbol\zeta)-(\mathbf X-\noise)\notag\\
 &\quad=\tilde f(\widetilde{\mathbf X}^s,s,\boldsymbol\zeta)+\Dnorm
   -(\widetilde{\mathbf X}-\noise)-\Dnorm\notag\\
 &\quad=\tilde f(\widetilde{\mathbf X}^s,s,\boldsymbol\zeta)
   -(\widetilde{\mathbf X}-\noise),
\end{align}
so the two fields have identical flow-matching regression errors on paired
paths, and hence identical losses.

\paragraph{Sampling.}
Let $\widetilde{\mathbf X}^s$ solve
$\tfrac{d}{ds}\widetilde{\mathbf X}^s
=\tilde f(\widetilde{\mathbf X}^s,s,\boldsymbol\zeta)$ and set
$\mathbf X^s=\widetilde{\mathbf X}^s+s\Dnorm$. Differentiating gives
$\tfrac{d}{ds}\mathbf X^s
=\tilde f(\mathbf X^s-s\Dnorm,s,\boldsymbol\zeta)+\Dnorm
=f(\mathbf X^s,s,\boldsymbol\zeta)$ with
$\mathbf X^0=\widetilde{\mathbf X}^0$. When the solutions exist and are
unique, the two ODEs therefore generate trajectories related by the same
translation, and their endpoints satisfy
$\mathbf X^1=\widetilde{\mathbf X}^1+\Dnorm$, which is the restoration of
\meq{eq:residual_to_raw_inference} in normalized units. The same relation
holds for the $K$-step Euler discretization of
\meq{eq:residual_euler}, step by step.

\paragraph{Scope.}
The network input $\mathbf z$ need not contain all of $\boldsymbol\zeta$:
current velocity alone does not identify the task-specific acceleration
estimate or the projection schedule. The correspondence then does not
assert equivalence between raw and residual fields conditioned only on
$\mathbf z$, although restoration remains algebraically exact for each
supplied displacement. Neither statement guarantees approximation accuracy,
equal expressiveness of a fixed architecture, or an optimization or success
gain for a finite network.

\subsection{Variation across velocities}
\label{supp:analysis_cov}
This subsection makes \meq{eq:covariance_difference} precise. Fix a context
$\boldsymbol\kappa$ that determines the observation, the task estimate
$\hat{\mathbf a}_\chi$, the action offsets $\tau_{1:H}$ and
$\tau_{\rm ref}$, the frame $\mathbf R$, and the centering reference
$\Dmean$; within this context the current velocity $\mathbf v=\mathbf v(t_0)$
still varies across demonstrations, because a single observation does not
determine it. Chunks are vectorized, second moments are finite, and
\begin{equation}
 \Sigma_{\mathbf v}=\operatorname{Cov}(\mathbf v\mid\boldsymbol\kappa),\qquad
 \mathbf w=(\tau_h-\tau_{\rm ref})_{h=1}^{H}\in\mathbb R^{H}.
 \label{eq:supp_cov_defs}
\end{equation}
Write $\Achunk=\mathbf B+\Dchunk$ with $\Dchunk=\Dchunk(\Vprof)$, so that
$\Ares=\mathbf B+\Dmean$, and let
$\mathbf C=\operatorname{Cov}(\mathbf B,\Dchunk\mid\boldsymbol\kappa)$.
$\mathbf B$ is the demonstrated behavior in the frame that moves with the
forecast and may depend on velocity. For symmetric matrices,
$\mathbf M\preceq\mathbf N$ means that $\mathbf N-\mathbf M$ is positive
semidefinite.

\begin{theorem}[Covariance of residual targets]
\label{thm:supp_cov}
Under the setting above:
\begin{enumerate}\itemsep1pt
\item[(i)] \emph{Identity.}
$\operatorname{Cov}(\Achunk\mid\boldsymbol\kappa)
-\operatorname{Cov}(\Ares\mid\boldsymbol\kappa)
=\operatorname{Cov}(\Dchunk\mid\boldsymbol\kappa)+\mathbf C+\mathbf C^\top$.
\item[(ii)] \emph{Removed variation.} On the translational channels,
$\operatorname{Cov}(\Dchunk\mid\boldsymbol\kappa)
=\mathbf w\mathbf w^\top\otimes\mathbf R\Sigma_{\mathbf v}\mathbf R^\top$,
and it is zero on the orientation and gripper channels; hence
$\operatorname{tr}\operatorname{Cov}(\Dchunk\mid\boldsymbol\kappa)
=\|\mathbf w\|^2\operatorname{tr}\Sigma_{\mathbf v}$.
\item[(iii)] \emph{Uncorrelated residual.} If $\mathbf C=\mathbf 0$, then
$\operatorname{Cov}(\Ares\mid\boldsymbol\kappa)\preceq
\operatorname{Cov}(\Achunk\mid\boldsymbol\kappa)$: no linear combination of
target channels has larger variance after residualization, and the total
variance decreases by exactly $\|\mathbf w\|^2\operatorname{tr}\Sigma_{\mathbf v}$,
which is positive whenever the velocity varies within the context and the
horizon extends beyond $\tau_{\rm ref}$.
\item[(iv)] \emph{Partial following.} If
$\Achunk=\mathbf B_\perp+\beta\Dchunk$ with
$\operatorname{Cov}(\mathbf B_\perp,\Dchunk\mid\boldsymbol\kappa)=\mathbf 0$
and a scalar $\beta$, i.e., the demonstrated chunk follows a fraction
$\beta$ of the forecast displacement, then
$\operatorname{Cov}(\Achunk\mid\boldsymbol\kappa)
-\operatorname{Cov}(\Ares\mid\boldsymbol\kappa)
=(2\beta-1)\operatorname{Cov}(\Dchunk\mid\boldsymbol\kappa)$. The covariance
decreases in the Loewner order if $\beta>\tfrac12$, is unchanged if
$\beta=\tfrac12$, and increases if $\beta<\tfrac12$.
\item[(v)] \emph{General sufficient condition.} The total variance decreases
whenever
$\operatorname{tr}\operatorname{Cov}(\Ares\mid\boldsymbol\kappa)
<\tfrac14\operatorname{tr}\operatorname{Cov}(\Dchunk\mid\boldsymbol\kappa)$.
\item[(vi)] \emph{Fixed forecast.}
$\operatorname{Cov}(\Ares\mid\boldsymbol\kappa,\Vprof)
=\operatorname{Cov}(\Achunk\mid\boldsymbol\kappa,\Vprof)$.
\end{enumerate}
\end{theorem}

\begin{proof}
(i) Given $\boldsymbol\kappa$, $\Dmean$ is constant, so
$\operatorname{Cov}(\Ares\mid\boldsymbol\kappa)
=\operatorname{Cov}(\mathbf B\mid\boldsymbol\kappa)$. Bilinearity of the
covariance gives
$\operatorname{Cov}(\mathbf B+\Dchunk\mid\boldsymbol\kappa)
=\operatorname{Cov}(\mathbf B\mid\boldsymbol\kappa)
+\operatorname{Cov}(\Dchunk\mid\boldsymbol\kappa)+\mathbf C+\mathbf C^\top$;
subtracting the two expressions proves (i).

(ii) By \meq{eq:forecast_displacement} and \meq{eq:kinematic_projection},
$\mathbf d_h=(\tau_h-\tau_{\rm ref})\mathbf R\mathbf v
+\tfrac12(\tau_h^2-\tau_{\rm ref}^2)\mathbf R\hat{\mathbf a}_\chi$, in which
only $\mathbf v$ is random given $\boldsymbol\kappa$. Hence
$\mathbf d_h-\mathbb E[\mathbf d_h\mid\boldsymbol\kappa]
=w_h\mathbf R(\mathbf v-\mathbb E[\mathbf v\mid\boldsymbol\kappa])$ and
$\operatorname{Cov}(\mathbf d_h,\mathbf d_{h'}\mid\boldsymbol\kappa)
=w_hw_{h'}\mathbf R\Sigma_{\mathbf v}\mathbf R^\top$, which is the stated
Kronecker form. The selector $\mathbf P$ in \meq{eq:displacement_chunk}
places zeros in the other channels. Taking the trace and using
$\mathbf R^\top\mathbf R=\mathbf I_3$ gives
$\sum_h w_h^2\operatorname{tr}\Sigma_{\mathbf v}$.

(iii) With $\mathbf C=\mathbf 0$, the difference in (i) is a covariance
matrix and therefore positive semidefinite. Its trace follows from (ii);
$\|\mathbf w\|^2>0$ when some $\tau_h\neq\tau_{\rm ref}$, and
$\operatorname{tr}\Sigma_{\mathbf v}>0$ when $\mathbf v$ is not constant
given $\boldsymbol\kappa$.

(iv) Here $\mathbf B=\Achunk-\Dchunk=\mathbf B_\perp+(\beta-1)\Dchunk$, so
$\mathbf C=(\beta-1)\operatorname{Cov}(\Dchunk\mid\boldsymbol\kappa)$, and
(i) gives $(1+2(\beta-1))\operatorname{Cov}(\Dchunk\mid\boldsymbol\kappa)$.

(v) Let $\sigma_B^2=\operatorname{tr}\operatorname{Cov}(\mathbf B\mid\boldsymbol\kappa)
=\operatorname{tr}\operatorname{Cov}(\Ares\mid\boldsymbol\kappa)$ and
$\sigma_D^2=\operatorname{tr}\operatorname{Cov}(\Dchunk\mid\boldsymbol\kappa)$.
Since $\operatorname{tr}\mathbf C=\mathbb E[\langle\mathbf B-\mathbb E\mathbf B,
\Dchunk-\mathbb E\Dchunk\rangle\mid\boldsymbol\kappa]$, the Cauchy--Schwarz
inequality gives $|\operatorname{tr}\mathbf C|\le\sigma_B\sigma_D$, so the
trace of the difference in (i) is at least
$\sigma_D^2-2\sigma_B\sigma_D=\sigma_D(\sigma_D-2\sigma_B)$, which is
positive when $\sigma_B^2<\tfrac14\sigma_D^2$.

(vi) $\Dchunk$ and $\Dmean$ are deterministic given $(\boldsymbol\kappa,\Vprof)$,
and adding a constant does not change a covariance.
\end{proof}

\paragraph{Interpretation and scope.}
Item (ii) shows that the variation KRA can remove grows with the spread of
object velocities within a context and with the sum of squared time offsets,
$\|\mathbf w\|^2$. With a fixed number of actions, stretching all offsets by
a factor multiplies this sum by the square of that factor. For the zero-based
projection with uniform spacing $\delta$,
$\|\mathbf w\|^2=\delta^2\sum_{h=1}^{H}(h-1)^2$. Items (iii) and
(iv) give the conditions under which this variation is actually removed:
residualization reduces target covariance when the demonstrations move with
the object, i.e., follow more than half of its forecast displacement, and
can increase it when they do not, for example during phases in which the
gripper is held still while the object passes. The theorem is therefore a
statement about the demonstrations, not a guarantee for every dataset. Item
(vi) shows that at a fixed forecast the spread is unchanged: residualization
changes how the targets vary \emph{across} velocities, not their spread at a
fixed velocity, and conditioning on current velocity alone fixes the
forecast only if it also identifies the task estimate and the projection
schedule. All statements concern the training targets of the conditional
flow; lower target covariance does not by itself guarantee a smaller
prediction error of a finite network, and the theorem makes no claim about
execution-time innovation.

\section{Ablation Protocols}
\label{supp:ablation}

\subsection{Component ablation}
\label{supp:components}
In each KRA configuration of Table~\mainref{tab:innovation_ablation}, the
forecast that constructs the residual training targets is the one used for
restoration, so the constant-velocity row is trained and restored with
$\hat{\mathbf a}_\chi=\mathbf 0$, and the VPD rows with the task estimates.
All four configurations use the current-velocity neural input and
DynamicVLA's native asynchronous streaming. The three KRA configurations
use a constant analytic gate of one. Disabling VI leaves the restored chunk $\Ahat$ unchanged and
removes only the offset $\mathbf c(t)$.

\paragraph{Significance tests.}
The Fisher exact tests of Sec.~\mainref{sec:exp_ablations} use the success
counts implied by the reported rates, $466$, $543$, $550$, and $554$ of
$1{,}800$ trials. They treat trials as independent and do not model the
pairing of scenes across configurations or variation across training
seeds. Matched controls that add the displacement to a raw-action policy
or use object-relative actions would be needed to attribute the residual
gain to a specific mechanism.

\subsection{Velocity measurement noise}
\label{supp:noise}
The Gaussian rows of Table~\mainref{tab:velocity_uncertainty} perturb
every velocity measurement as
\begin{equation}
 \widetilde{\mathbf v}_k=\mathbf v_k+\boldsymbol\epsilon_k,
 \qquad\boldsymbol\epsilon_k\sim
 \mathcal N(\mathbf0,\sigma_v^2\mathbf I_3),
\end{equation}
for each measurement $k$. The same corrupted measurement feeds
the neural input, the planning-time forecast, and VI wherever they are
enabled, so the baseline and KRA are perturbed in the same way.
Normalization statistics, the centering reference, simulator motion, and
task scoring are unchanged. The Gaussian perturbations isolate random,
zero-mean measurement error.

\paragraph{EfficientTAM-Ti velocity.}
In the last row of the table, simulator-state velocity is replaced, for
all three policies and wherever velocity is used, by a causal estimate from
EfficientTAM-Ti masks in the simulated RGB-D observations. At the first
frame, the simulator supplies a ground-truth instance mask for each initially
visible candidate object. EfficientTAM-Ti propagates these masks from that
single initialization. The benchmark supplies the active target identity
throughout the rollout. When the target changes, that identity selects the
corresponding track if the object was seeded in the first frame.

The velocity estimator fits a causal slope to each 3D position component
over the preceding $0.16$\,s. If a missing mask or invalid depth prevents a
3D position measurement, the estimator clears that track's velocity history.
It resumes estimation after observing at least three fresh valid positions.
Whenever the active target has no valid velocity estimate, the policy receives
$\mathbf{0}$\,m/s. This fallback also applies whenever an object without an
initial mask becomes the active target.

\ifdefined\includetrackingdetails
\begin{figure*}[t]
  \centering
  \begin{lrbox}{\paperfigurebox}\begin{minipage}{\textwidth}
  \centering
  \includegraphics[width=\textwidth,height=.24\textheight,keepaspectratio]{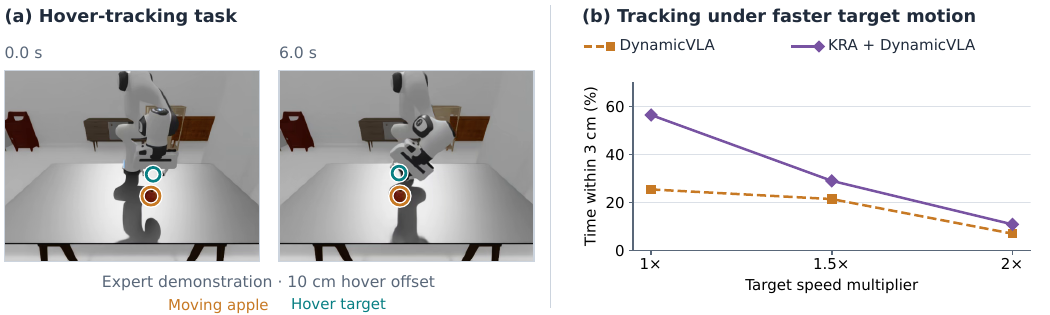}
  \caption{\textbf{Hover tracking.} (a)~Expert demonstration frames with the projected apple center and the hover target 10\,cm above it. (b)~Coverage of the velocity-conditioned DynamicVLA and KRA + DynamicVLA on the ten held-out paths of the speed sweep (Sec.~\mainref{sec:exp_tracking}), replayed at $1\times$, $1.5\times$, and $2\times$ the reference speed.}
  \label{fig:tracking_evidence}
  \end{minipage}\end{lrbox}
  \checkpaperfigure{tracking}
\end{figure*}

\fi

\section{Controlled Hover Tracking Protocol}
\label{supp:tracking}
\label{app:tracking_details}

\paragraph{Task.}
The robot follows a point 10\,cm above the visual top of a moving apple
without making contact (Figure~\ref{fig:tracking_evidence}a). The four
DynamicVLA variants of Table~\mainref{tab:tracking} cross raw versus residual
action coordinates with the presence or absence of neural velocity input.
Tracking LS is sustained following, not a multi-object manipulation
sequence.

\paragraph{Demonstrations and training.}
The 300 expert demonstrations combine object-velocity feedforward with
Cartesian position--velocity feedback mapped to joint commands by damped
inverse kinematics. Each records scene and wrist RGB at 25\,Hz, the
instruction, proprioception, and the simulator object velocity, split
80/10/10 into training, validation, and test episodes. All four variants
share a 75-epoch recipe with a 20-action horizon at $0.04$\,s spacing,
training-split normalization, and validation-based checkpoint selection.

\paragraph{Evaluation.}
Physics and scoring run at 200\,Hz. After a 3\,s expert warm-up, the policy
replans every 20 camera frames ($0.8$\,s) with physics paused during
inference. The two KRA variants restore the displacement across each
$0.8$\,s window using simulator velocity. All four variants disable VI and
asynchronous streaming. CR, DA, and LS run 200 matched trials per variant, 2,400 rollouts
in total. Coverage counts the whole episode, including the first scored
second, and a policy that terminates early receives no credit for the
unexecuted remainder of its requested horizon. All trials of
Table~\mainref{tab:tracking} complete their requested horizons. RMSE is the
mean of per-episode 3D position RMSE.

\paragraph{Speed sweep.}
The speed sweep of Sec.~\mainref{sec:exp_tracking} uses ten paths held out
from training. Each of the two velocity-conditioned policies runs once on
each path at each speed, giving 20 trials across both policies per speed.
The paths are replayed at the reference speed and at $1.5\times$ and
$2\times$ that speed. Faster replay increases both velocity and acceleration
along the same path. Coverage uses the same 3\,cm threshold and whole-episode
scoring as above. Figure~\ref{fig:tracking_evidence}b shows coverage at each
speed. At $2\times$, all trials for both policies finish. KRA + DynamicVLA
has higher coverage (10.89\% vs.\ 7.02\%), and DynamicVLA has lower position
RMSE (72.50 vs.\ 81.78\,mm).

\section{Manipulation and Real-Robot Details}
\label{supp:exp_details}

\subsection{Low-clutter DOM}
\label{supp:exp_dom}
\paragraph{Benchmark construction.}
The low-clutter suite uses DOM~\cite{xie2026dynamicvla} scenes with at most
two movable task objects and at most five objects in total, including
destination containers and static objects. We sort the movable rigid-object
keys by name, retain the primary object and the first named distractor when
present, and remove the remaining movable objects. All destination containers
are retained. This changes 18 of the 90 scenes. The resulting suite has 44
scenes with one movable object and 46 with two.

\paragraph{Matched trials.}
All checkpoints in Table~\mainref{tab:dom_main} share scene configurations,
reset seeds, initial poses and velocities, and a frozen schedule of 20 prompt
variants per scene. The adaptation preserves the physics of retained objects,
camera settings, forces, and episode horizons. We regenerate descriptions for
capped placement scenes from the retained geometry. Long-horizon scenes keep
the ``entire set'' instruction and use the trimmed target list. Evaluations run
in Isaac Sim 4.5.0 with Isaac Lab 2.2.1. The 80 placement scenes have a 12\,s
horizon and the ten long-horizon scenes have a 20\,s horizon. The 20 repeats per
scene yield 200 trials per condition and 1,800 trials per checkpoint.

\paragraph{Success criterion.}
All configured target objects must pass the placement test and the end
effector must return within 0.07\,m of its configured goal. An object passes
when its center lies within the padded destination footprint and its lowest
point lies at or below the destination's upper bound. The scorer checks
this geometry at a single timestep, so a transient placement can pass even
with an unsupported object or an unverified grasp or release. In the recorded
rollout videos, some logged successes show the object on the destination
rim rather than inside it.
Overall success averages the nine equally sized conditions. Mean termination
time includes successes, failures, and timeouts.

\paragraph{Fine-tuning.}
The common 20,000-episode fine-tuning subset contains 9,000 pick, 9,000
place, and 2,000 long-horizon episodes. Selection is stratified by task and
object speed, keeps duplicate groups together, and preserves the original
held-out split. Table~\ref{tab:supp_training_recipes} lists the planned
budgets for this fine-tuning stage. The fresh DynamicVLA recipe starts from
the released pretrained weights. The SmolVLA, VLASH, and $\pi_{0.5}$ recipes
continue from earlier training checkpoints. The recorded recipes use AdamW
with $\beta=(0.9,0.95)$, $\epsilon=10^{-8}$, 1,000 warm-up updates and a cosine
schedule. The adapter schedules have a floor of $2.5\times10^{-6}$.
Continuation on the 20,000-episode subset restores the optimizer and parent
schedule, whose decay can end before the final training update. We monitor
held-out mean squared error of reconstructed 20-action chunks in physical units.

\begin{table}[t]
  \centering
  \caption{Recorded low-clutter DOM training recipes. The batch size is
  global. Update budgets are for ten epochs on the 20,000-episode subset.
  LR denotes the configured peak learning rate.}
  \label{tab:supp_training_recipes}
  \small
  \setlength{\tabcolsep}{4pt}
  \begin{tabular}{lrrrr}
    \toprule
    Backbone & Batch & Updates & Peak LR & Weight decay \\
    \midrule
    DynamicVLA & 352 & 68,380 & $3\times10^{-5}$ & $10^{-10}$ \\
    SmolVLA & 64 & 376,050 & $10^{-4}$ & $10^{-10}$ \\
    VLASH & 32 & 752,110 & $5\times10^{-5}$ & $10^{-10}$ \\
    $\pi_{0.5}$ & 32 & 752,110 & $2.5\times10^{-5}$ & $10^{-2}$ \\
    \bottomrule
  \end{tabular}
\end{table}

\paragraph{Trainable modules.}
DynamicVLA updates the vision encoder, connector, language model, action
expert, and state projection. SmolVLA updates its action expert and state
projection with a frozen vision encoder. VLASH updates its vision encoder,
connector, language model, action expert, and projections. All 3.62 billion
parameters are eligible for updates in our VLASH configuration.
The $\pi_{0.5}$ adapter uses native fine-tuning with gradient checkpointing
and adds no parameter freezes.

\paragraph{Observation and action timing.}
DynamicVLA uses frames at offsets $-2$ and $0$ ($-80$ and $0$\,ms).
SmolVLA and $\pi_{0.5}$ use current front and wrist images. These three
backbones use action-label offsets $0$ through $19$. VLASH uses current
images and samples a delay $d\in\{0,1,2,3\}$ frames during training, with
action labels at offsets $d$ through $d+19$. All four recipes space action
labels by 40\,ms (25\,Hz), so the first and last labels of a chunk are
760\,ms apart.

\subsection{Real robot}
\label{supp:exp_real}
\paragraph{Data.}
The 900 demonstrations comprise 300 per object, collected across two box
positions and two entry regions and including stationary grasps. Both
policies are fine-tuned on the same 700-demonstration training partition and
receive the EfficientTAM velocity of Sec.~\ref{supp:forecast} as their
neural input. KRA's projection uses the fixed deceleration estimate for
the object and ramp height of the trial.

\paragraph{Trials and counts.}
In terms of counts, Table~\mainref{tab:real_robot_results} corresponds to
$10\to13$ (orange), $7\to8$ (ball), and $5\to7$ (marker) successes out of
20 at $v\approx0.3$\,m/s, and $8\to10$, $5\to7$, and $4\to5$ at
$v\approx0.5$\,m/s, for DynamicVLA and KRA + DynamicVLA. The Fisher exact
test of Sec.~\mainref{sec:exp_real} treats these as independent trials and
does not model paired initial conditions or heterogeneity between cells.

\section{Computational Cost}
\label{supp:runtime}
\label{app:runtime_scope}
KRA leaves the backbone's network, parameter count, and number of network
evaluations per chunk unchanged; its additions are arithmetic on
quantities available at planning time or at the control rate.
Per planned chunk, the projection of Algorithm~\ref{alg:supp_project} and
the restoration of \meq{eq:residual_to_raw_inference} cost $O(H)$ vector
operations on the three translational channels, for example $H=20$ in hover
tracking. Per control step, VI updates one prefix integral and evaluates
two Euclidean-ball projections (Algorithm~\ref{alg:supp_vi}), with no
additional policy inference. Training adds the same projection and one
subtraction per sampled chunk, and the reference $\Dmean$ is computed once
in a single pass over the training chunks. These counts exclude the velocity source. The primary simulation
experiments read simulator state. The RGB-D condition in
Table~\mainref{tab:velocity_uncertainty} and the real-robot experiment
each use a tracker for both the baseline and KRA. We do not report a controlled
end-to-end latency comparison, so no claim about deployment overhead is
made beyond this operation count.

\section{Further Limitations}
\label{supp:limitations}
This section adds to the limitations stated in the main paper.

\paragraph{Method.}
The constant gate keeps VI active through gripper closing and contact, where
a translation offset can disturb a grasp that is being established; a
phase-dependent gate was not evaluated. The forecast uses one fixed
acceleration estimate for a given setup, so motion that changes regime
within a chunk, such as the transition from ramp to table in
Figure~\ref{fig:supp_ramp_velocity}, is left to VI and the learned residual.
The residual-flow correspondence and the covariance identity of
Sec.~\ref{supp:analysis} hold for unrestricted fields and do not guarantee an
optimization or sample-efficiency benefit for a finite network.

\paragraph{Evaluation.}
In simulation the velocity input belongs to the designated target object;
in the multi-object perception conditions (VU, SR, MP) it can therefore
also help identify the target. The EfficientTAM-Ti study also uses the
simulator's initial masks and active target identity
(Sec.~\ref{supp:noise}). The low-clutter protocol and the evaluated
checkpoints define the scope of the cross-backbone results. The physical
study tests closed-loop pick-and-place from a ramp, not the full set of DOM
conditions, and does not report failure counts by phase (interception,
grasp, transport, placement). The increments from VPD and VI are not
statistically established under the independent-trial analysis; paired
analysis over matched scenes, repeated training runs, and direct
measurements of target variation across velocities would strengthen the
attribution of the observed improvements.

\section{Supplementary Video}
\label{supp:video}
The video presents Kinematic Residual Adaptation (KRA) and selected
closed-loop rollouts. All four baseline VLAs shown receive the current
3D object velocity as a neural input. Success and failure labels appear
when the outcome becomes visible, or at trial end if the outcome remains
ambiguous. The main-paper tables report aggregate results.

\paragraph{Method.}
The opening animations follow a single action chunk through planning,
training, inference, and execution (Sec.~\mainref{sec:method}).
Velocity-profile displacement (VPD) forecasts the object's translation
at each scheduled action. The policy learns a residual after subtraction
of the centered displacement. At inference, KRA restores that displacement.
During execution, velocity innovation (VI) applies a bounded translation
offset based on the difference between measured and forecast motion.

\paragraph{Low-clutter DOM.}
The columns show DynamicVLA, $\pi_{0.5}$, VLASH, and SmolVLA from left
to right. Each upper clip shows the velocity-conditioned baseline.
The purple \textbf{+KRA} label below it identifies the same backbone
with KRA. The three scenes illustrate closed-loop reactivity (avocado
into a yellow box), motion perception (initially faster object into a red
box), and motion generalization (avocado into a red box). Each scene is
shared across eight archived clips selected to show baseline misses and
placements with KRA. Selected repeats and prompt wording can
differ across policies. Visible placements and misses agree with the
logged outcomes in these examples.

\paragraph{Ablation Study 1.}
Two rollouts share a KRA checkpoint, VPD forecast, scene, reset seed,
prompt, and repeat index. With VPD alone, the robot misses the avocado.
With VI enabled, it places the avocado in the red box.
Table~\mainref{tab:innovation_ablation} gives the aggregate component ablation.

\paragraph{Ablation Study 2 and tracker evaluation.}
At per-axis Gaussian noise levels of $0.05$ and $0.10$\,m/s, each row
pairs a baseline miss with a KRA placement from another repeat of the
same scene. These examples and the simulated EfficientTAM-Ti clips use
a separate constant-velocity KRA evaluation with VI disabled. Both KRA
columns of Table~\mainref{tab:velocity_uncertainty} use VPD.
Because the real robot estimates velocity from masks and depth, we also
compare the tracker against simulator ground truth. A matched pair shows
the lemon-and-tray task. The potato rollout plots the original logged
velocities. Its cyan outline was reconstructed from a manually traced
first-frame outline for the replay. Sec.~\ref{supp:noise} describes the
simulator-provided masks used during evaluation.

\paragraph{Real robot and future work.}
On the ALOHA VX300 arm, KRA places the orange. For the striped ball and
marker, each pair shows separate releases under similar conditions.
Table~\mainref{tab:real_robot_results} gives completion rates for all
objects and speeds. The penultimate scene shows a matched simulated trial
at $0.01$\,m/s Gaussian noise per axis. DynamicVLA places the lemon in
the tray. Constant-velocity KRA grasps it but misses during placement.
The episode motivates studying when to enable motion compensation.
The conclusion returns to the approach diagram in Fig.~2 of the main paper.

\end{document}